\documentclass[letterpaper]{article}

\usepackage[preprint]{aaai2027}
\usepackage[hyphens]{url}
\usepackage{graphicx}
\usepackage{natbib}
\usepackage{caption}
\usepackage{booktabs}
\usepackage{multirow}
\usepackage{xcolor}
\usepackage{colortbl}
\usepackage{arydshln}
\usepackage{adjustbox}
\usepackage{amsmath,amssymb}

\usepackage{booktabs}
\usepackage{multirow}
\usepackage{pifont}
\usepackage{xcolor}

\usepackage{pgfplots}
\definecolor{projblue}{HTML}{2563EB}
\definecolor{softblack}{HTML}{333333}

\usepackage{tcolorbox} \tcbuselibrary{breakable,skins} \newtcolorbox{promptbox}[1]{ enhanced, breakable, colframe=black, colback=white, coltitle=white, colbacktitle=black, title={#1}, fonttitle=\small\bfseries, fontupper=\small, boxrule=0.8pt, arc=1.5mm, outer arc=1.5mm, left=4mm, right=4mm, top=3mm, bottom=3mm, before skip=10pt, after skip=12pt, before upper={ \setlength{\parindent}{0pt} \setlength{\parskip}{0.5em} } }

\definecolor{keepgreen}{RGB}{0,150,85}
\definecolor{keepred}{RGB}{210,45,45}

\DeclareRobustCommand{\keepyes}{%
  \textcolor{keepgreen}{\ding{51}}%
}
\DeclareRobustCommand{\keepno}{%
  \textcolor{keepred}{\ding{55}}%
}

\newcommand{\MODELNAME}{\textsc{KEEP-Bench}}

\title{\MODELNAME{}: Evaluating Evidence Preservation in Context Compression}

\title{\MODELNAME{}: Evaluating Evidence Preservation in Context Compression}

\author{
    Ziqing Man,
    Di Wu,
    Jia-Chen Gu,
    Junyi Zhang,
    Kai-Wei Chang,
    Nanyun Peng
}
\affiliations{
    University of California, Los Angeles\\
    \{mansionzq@cs., diwu@cs., gujc@, junyizhang2002@, kwchang@cs., violetpeng@cs.\}ucla.edu
}

\begin{document}

\maketitle

\begin{abstract}
Context compression improves long-context efficiency but may introduce faithfulness issues through an information bottleneck.
Existing evaluations rely on QA metrics, but a reader can still answer correctly from an unfaithful compression, so these metrics underestimate such loss.
To bridge this gap, we introduce \MODELNAME{}, a diagnostic benchmark that directly assesses compression faithfulness with 4,102 cases across three challenging settings that require keeping evidence for \emph{plausible reference resolutions}, \emph{explicit qualifiers}, and \emph{minority perspectives}.
Experiments show that even with tailored prompting, the strongest models remain unreliable: despite context windows exceeding 100K tokens and inputs averaging only 3.1K tokens, the best open-weight model (31B) preserves all required evidence in only 66.8\% of cases, and the best proprietary model in only 61.6\%.
Performance also varies sharply across evidence structures, so conclusions from limited structures may be biased.
We further introduce a metric, Evidence Preservation Return (EPR), to relate evidence recall to inference FLOPs, and find that it peaks at moderate scales (4B--8B), where recall remains low.
Additional output budget mainly benefits already strong models and widens capability gaps, yet the stronger preservation of these models comes at a steep computational cost.
These findings highlight the importance for future work to investigate compressors with strong information preservation under a small computation budget, rather than relying on a large output budget or model scaling.
\end{abstract}

\begin{links}
    \link{Code}{https://github.com/PlusLabNLP/KEEP-Bench}
    \link{Datasets}{https://hf.co/datasets/uclanlp/KEEP-Bench}
    \link{Website}{https://pluslabnlp.github.io/KEEP-Bench/}
\end{links}

\section{Introduction}
Context compression aims to accelerate inference and reduce the cost of long-context processing in large language models (LLMs) without compromising downstream reasoning~\citep{li2023compressingcontextenhanceinference,jiang-etal-2023-llmlingua,jiang-etal-2024-longllmlingua,ge2024incontextautoencodercontextcompression}.
Yet compression simultaneously creates an information bottleneck: omitting or distorting critical evidence can render the compressed context unfaithful~\citep{pan-etal-2024-llmlingua,lajewska-etal-2025-understanding}.
In a controlled pilot study, we compare downstream accuracy using original, faithfully compressed, and unfaithfully compressed contexts.
Accuracy on faithful contexts is comparable to that on original contexts, whereas accuracy on unfaithful contexts is over 80 percentage points lower.
This gap highlights the consequences of evidence loss and underscores the importance of accurately evaluating compression faithfulness.
See Supplementary Section~\ref{app:downstream_pilot} for details.

\begin{figure*}[!t]
    \centering
    \includegraphics[width=\textwidth]
    {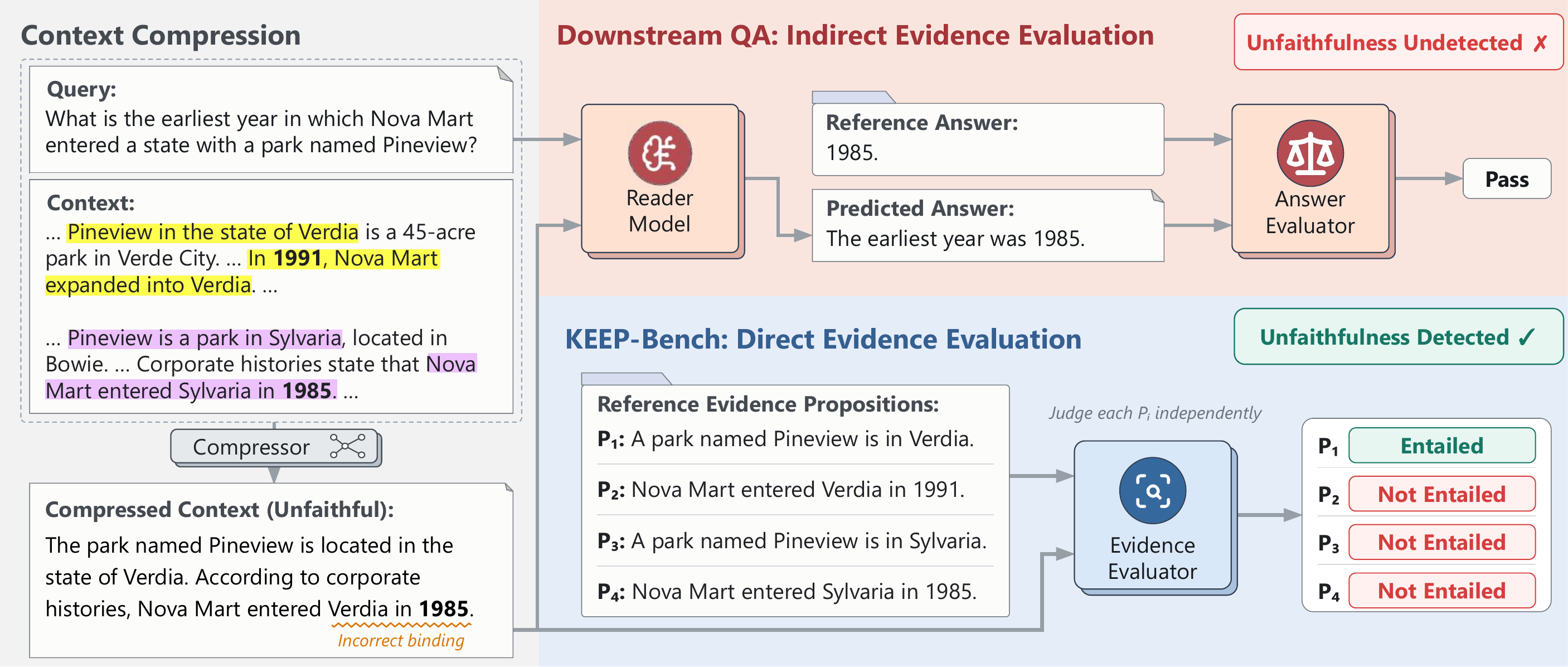}
    \caption{
    Comparison of indirect evaluation and direct evidence evaluation on a simplified \MODELNAME{} case. The upper branch shows indirect evaluation through downstream QA, which considers only the final answer and may therefore falsely accept an unfaithful compressed context. The lower branch shows \MODELNAME{}’s direct evidence evaluation, which independently checks whether each reference evidence proposition is entailed by the compressed context, thereby detecting the preservation failure.
    }
    \label{fig:teaser}
\end{figure*}
Despite its importance, compression faithfulness remains inadequately evaluated.
Two limitations of existing evaluation protocols obscure faithfulness failures.
First, compression quality is typically inferred from downstream task performance~\citep{jiang-etal-2024-longllmlingua,yoon-etal-2024-compact,li-etal-2025-brief}. 
As shown in Figure~\ref{fig:teaser}, this proxy can be misleading, as an unfaithful compressed context may lead to a correct answer through the reader's parametric knowledge or task shortcuts.
Second, existing compressor evaluations often rely on QA cases in which the answer is supported by a single, relatively direct reasoning path~\citep{lee-etal-2025-grade,zhu-etal-2024-towards-graph}.
In such cases, the required evidence is comparatively easy for compressors to identify, leaving preservation under more complex evidence structures largely untested.

To address these limitations, we introduce \MODELNAME{} (\textbf{K}ey \textbf{E}videnc\textbf{E} \textbf{P}reservation), a benchmark designed to directly evaluate whether context compression faithfully preserves the evidence required for downstream reasoning.
As illustrated in the lower branch of Figure~\ref{fig:teaser}, \MODELNAME{} assesses the compressed context itself rather than downstream task performance, isolating compression fidelity from reader behavior and exposing evidence loss that answer accuracy may otherwise conceal.
\MODELNAME{} comprises 4,102 cases across three diagnostic settings, each targeting a distinct evidence structure.
First, the \emph{parallel evidence} setting contains underspecified entity mentions that may correspond to two distinct referents, inducing two parallel reasoning chains that must both be preserved.
Second, the \emph{conditional evidence} setting requires retaining both supporting evidence and its crucial accompanying qualifiers, as omitting either misrepresents the compressed context.
Third, the \emph{imbalanced evidence} setting presents multiple valid perspectives supported by unequal amounts of evidence, testing whether compression also captures the minority perspectives.

To measure evidence preservation directly, \textsc{KEEP-Bench} extracts all evidence necessary for each query from the original context and decomposes it into a set of minimal, self-contained propositions.
It then reports \emph{All Proposition Recall (All-Rec)}, which credits an instance only when every proposition is preserved in the resulting compressed context.
We further introduce \emph{Evidence Preservation Return} (EPR), derived from the ratio of overall All-Rec to average inference TFLOPs per instance, to quantify how effectively a compressor converts computation into preservation.
We evaluate seventeen compressor configurations, including dedicated compressors and general LLMs, across diverse evidence structures.
All dedicated compressors remain below 10\% overall All-Rec, and general LLMs under 10B parameters remain below 30\%.
Stronger general LLMs perform better, yet Gemma-4-31B-it and Claude-Haiku-4.5 achieve only 66.8\% and 61.6\% overall All-Rec, respectively, although source contexts average only 3.1K tokens, far below the evaluated LLMs' 32K--400K context windows.
Performance also varies sharply across settings: Claude-Haiku-4.5 leads the parallel setting with twice the All-Rec of GPT-5.4-mini, yet trails it in imbalanced science cases, so evaluation on a single evidence structure can misjudge compressors.
EPR further shows that scaling improves preservation but not always in proportion to the added computation, with moderate scales (4B--8B) offering the best balance within each family.

\vspace{-1mm}

Further analyses characterize evidence preservation along three dimensions: computational efficiency, output budget, and evidence position.
First, our analysis reveals a substantial efficiency gap: approaching the preservation performance of large models at relatively lower computational cost remains an open challenge.
Second, additional output budget mainly benefits already strong models, widening rather than closing capability gaps, and helps only within evidence structures a compressor already recognizes.
Third, greater separation between related pieces of evidence reduces their joint preservation, while evidence appearing earlier in the context is more likely to be retained.
Together, these findings call for compressors that better recognize evidence structures and achieve strong preservation with less computation.

\begin{table*}[t]
\centering

\begingroup
\small
\rmfamily
\setlength{\tabcolsep}{3pt}
\renewcommand{\arraystretch}{1.06}

\begin{tabular*}{\textwidth}{
  @{\extracolsep{\fill}}lcc*{5}{c}@{}
}
\toprule

\multirow{2}{*}{\bfseries Benchmark}
& \multirow{2}{*}{\bfseries Direct Eval.}
& \multirow{2}{*}{\bfseries Cases}
& \multicolumn{5}{c}{\bfseries Challenges} \\

\cmidrule(lr){4-8}

& & &
Multi-hop
& Fictional
& Parallel
& Condition
& Imbalance \\

\midrule

MuSiQue-Ans~\citep{trivedi-etal-2022-musique}
& \keepno
& 24,814
& \keepyes
& \keepno
& \keepno
& \keepno
& \keepno \\

ConditionalQA~\citep{sun-etal-2022-conditionalqa}
& \keepno
& 3,427
& \keepyes
& \keepno
& \keepno
& \keepyes
& \keepno \\

CompressionBench~\citep{zakazov2025cmprsr}
& \keepno
& 400
& \keepyes
& \keepno
& \keepno
& \keepno
& \keepno \\

Information Preservation Eval.~\citep{lajewska-etal-2025-understanding}
& \keepyes
& 900
& \keepyes
& \keepno
& \keepno
& \keepno
& \keepno \\

DeepAmbigQA~\citep{ji2025deepambigqaambiguousmultihopquestions}
& \keepno
& 3,600
& \keepyes
& \keepno
& \keepyes
& \keepno
& \keepno \\

Context Codec~\citep{trukhina2026contextcodec}
& \keepyes
& 5
& \keepno
& \keepno
& \keepno
& \keepno
& \keepno \\

MARCH~\citep{park-etal-2026-march}
& \keepno
& 2,209
& \keepyes
& \keepno
& \keepyes
& \keepno
& \keepno \\

DebateQA~\citep{xu-etal-2026-debateqa}
& \keepno
& 2,941
& \keepno
& \keepno
& \keepno
& \keepno
& \keepno \\

GroupQA~\citep{naphade-2026-rational}
& \keepno
& 1,635
& \keepno
& \keepno
& \keepno
& \keepno
& \keepyes \\

\midrule

\textbf{\MODELNAME{} (Ours)}
& \keepyes
& \textbf{4,102}
& \keepyes
& \keepyes
& \keepyes
& \keepyes
& \keepyes \\

\bottomrule
\end{tabular*}
\endgroup

\caption{
Comparison of \MODELNAME{} with representative benchmarks. 
\textit{Direct Eval.} indicates whether the compressed context is evaluated directly, rather than only through downstream task performance. 
\keepyes~and \keepno~indicate whether a property is present in at least part of a benchmark.
\textit{Fictional} denotes systematic entity fictionalization.
\textit{Parallel}, \textit{Condition}, and \textit{Imbalance} correspond to the three diagnostic settings described in the main text.
\textit{Cases} reports the number of evaluation instances per benchmark.
}
\label{tab:benchmark-comparison}
\end{table*}

\section{Related Work}

\paragraph{Context Compression}
Context compression is widely studied in RAG and increasingly applied to long-horizon agent contexts.
Existing systems instantiate compressors in several ways.
One pathway introduces specialized compressors, implemented as trained language models or encoder-based selectors, as in RECOMP~\citep{xu2024recomp}, BRIEF-Pro~\citep{gu-etal-2026-brief}, EXIT~\citep{hwang-etal-2025-exit}, and LLMLingua-2~\citep{pan-etal-2024-llmlingua}.
Another pathway prompts general small language models to generate textual compressions, sometimes with task-specific demonstrations or further optimization~\citep{pu2024stylecompress,zakazov2025cmprsr}.
Some systems instead use backbone LLMs to refine retrieved documents or accumulated interaction histories, requiring extra inference~\citep{li2025searcho1,kang2025acon}.
Finally, some work compresses context into latent representations~\citep{cheng2024xrag,ge2024incontextautoencodercontextcompression}, but we focus on textual compression because its outputs can be directly inspected for evidence preservation at the proposition level.

\paragraph{Compressor Evaluation}
Compressors are commonly evaluated indirectly through the downstream QA performance of a reader given the compressed context~\citep{li2023compressingcontextenhanceinference,jiang-etal-2023-llmlingua,jiang-etal-2024-longllmlingua,yoon-etal-2024-compact,li-etal-2025-brief,zakazov2025cmprsr}.
This conflates compression faithfulness with the task setup and reader behavior, while necessary evidence in conventional QA datasets is typically easy to identify.
Recent work evaluates compressed contexts more directly through response grounding and context reconstruction~\citep{lajewska-etal-2025-understanding}, citation grounding~\citep{li-etal-2026-efficiency}, evidential sufficiency~\citep{jeong-etal-2025-ecorag}, and atom-level commitment survival~\citep{trukhina2026contextcodec}.
\MODELNAME{} complements these approaches with controlled RAG-style cases that directly evaluate whether necessary propositions survive compression under parallel, conditional, and imbalanced evidence structures.

\paragraph{Evidence Structures in QA}
Prior QA benchmarks consider evidence structures that partially overlap with those in \MODELNAME{}.
However, existing work formulates these challenges as QA tasks and evaluates answer quality rather than compression performance.
DeepAmbigQA~\citep{ji2025deepambigqaambiguousmultihopquestions} and MARCH~\citep{park-etal-2026-march} evaluate answer completeness and reasoning under multi-hop ambiguity; \MODELNAME{} instead tests whether compression preserves both parallel evidence chains induced by an underspecified entity mention.
ConditionalQA~\citep{sun-etal-2022-conditionalqa} considers questions whose answers may differ across conditions; \MODELNAME{} instead fixes a single answer and reasoning chain in which one evidence proposition is valid only under a specific condition that compression must preserve.
For multiple perspectives, DebateQA~\citep{xu-etal-2026-debateqa} evaluates perspective coverage in generated answers, while GroupQA~\citep{naphade-2026-rational} studies how repeated or conflicting evidence affects RAG answers; \MODELNAME{} instead controls how much context supports each perspective and measures their preservation after compression.

\section{\MODELNAME{}}
\label{sec:benchmark}

\subsection{Task Formulation}
\label{sec:task_formulation} 
Given a query $q$ and a source context ${c}$, a compressor produces a compressed context $\tilde{c}$ from $c$. 
\MODELNAME{} evaluates evidence preservation by assessing whether $\tilde{c}$ faithfully preserves all evidence in ${c}$ required to carry out the reasoning demanded by $q$.
For each instance, we represent this evidence as a set of evidence propositions, denoted by $\mathcal{E}_q(c)=\{p_1,\ldots,p_n\}$.
Each $p_i$ is self-contained, allowing its preservation in $\tilde{c}$ to be assessed independently of the others. %
    
A compressed context is considered faithful if every evidence proposition is preserved.
An evidence proposition $p_i$ is considered preserved if it is stated in, or unambiguously inferable from, $\tilde{c}$ alone.
If $\tilde{c}$ omits $p_i$ or states information inconsistent with it, $p_i$ is considered not preserved. %

\subsection{Diagnostic Settings}

\MODELNAME{} comprises three distinct diagnostic settings, each designed to challenge a common assumption in existing evaluations. 
Table~\ref{tab:benchmark-comparison} situates \MODELNAME{} among representative compression and QA benchmarks, while Figure~\ref{fig:keep-bench-examples} shows representative examples of the three settings.

\subsubsection{Parallel Evidence}
The parallel evidence setting is built on fictional factoid-style multi-hop questions.
All evidence entities are fictional, and the required evidence is stated only in the context, preventing compressors from recovering omitted evidence from memorized world knowledge.
Each instance contains one or more underspecified entity mentions whose referents differ between two valid interpretations.
Each interpretation induces a separate reasoning chain.
The two chains are parallel in logical structure but involve disjoint referents and evidence.
Each reasoning chain consists of one evidence proposition per reasoning hop, and the complete evidence proposition set is the union of both chains.
The benchmark includes two-, three-, and four-hop cases.

\subsubsection{Conditional Evidence}
The conditional evidence setting shares the same synthetic setup as the parallel evidence setting, except that each instance contains a single reasoning chain.
One reasoning step is accompanied by an explicit qualifier that defines the condition under which it holds; the context also states that the proposition does not hold when the condition is unmet.
Accordingly, for each conditional instance, the evidence proposition set consists of one proposition per reasoning hop together with the required qualifier.

\subsubsection{Imbalanced Evidence}
The imbalanced evidence setting consists of open-ended queries.
Each evidence proposition in the context addresses the query from a distinct angle.
Dominant propositions receive extensive supporting text, whereas each minority proposition, though equally required, appears in only a single sentence.
When multiple propositions are dominant, their support is kept roughly balanced.
Together, all such propositions in the given context constitute the complete evidence proposition set for each case.
The setting covers two application domains: science queries concerning broad topics explored across multiple studies, and policy queries concerning rules that vary across regions in their requirements, deadlines, or applicability.

\begin{figure}[t]
    \centering
    \includegraphics[width=\columnwidth]{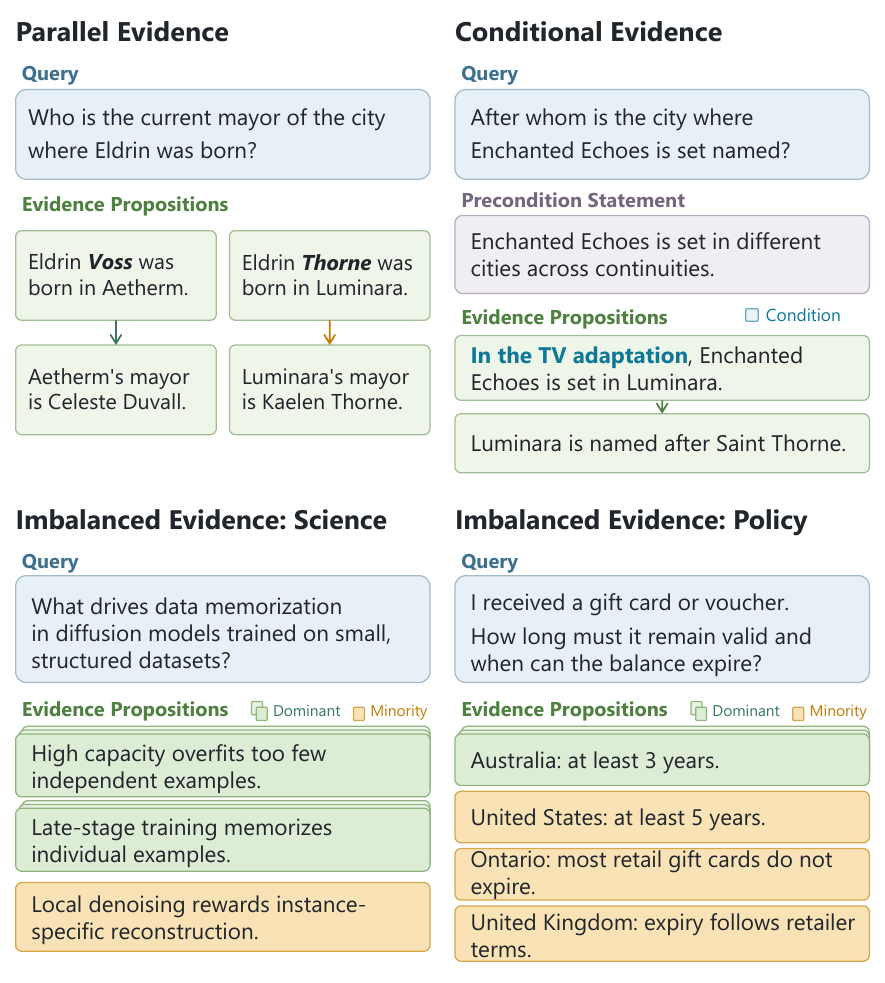}
    \caption{Illustrative examples of the diagnostic settings in \MODELNAME{}.
    Parallel evidence requires preserving evidence for two valid interpretations;
    conditional evidence requires preserving an explicit condition;
    and imbalanced evidence requires preserving dominant and minority evidence.}
    \label{fig:keep-bench-examples}
\end{figure}

\subsection{Data Construction}
\label{sec:data_construction}

To create cases for both parallel and conditional evidence settings, we first select qualified MuSiQue~\citep{trivedi-etal-2022-musique} examples, extract one minimal evidence proposition to represent each reasoning hop, and fictionalize the entities in the query and evidence propositions.
The two settings then differ only in how the reasoning chains are transformed.

\paragraph{Parallel Evidence.}
We replace one or more queried entities with underspecified fictional mentions.
For one such mention $r$, we introduce two distinct referents, $r_1$ and $r_2$, each associated with one separate but parallel reasoning chain, such that $r$ may refer to either.
We then write separate supporting paragraphs for the two chains and fictionalize the distractor paragraphs to ensure consistency throughout the context.

\paragraph{Conditional Evidence.}

Starting from the same fictionalized reasoning chain, we add an explicit qualifier specifying when one particular reasoning hop is valid.
We then write supporting paragraphs for the conditioned chain together with a separate disambiguating paragraph stating that the selected relation does not hold when the condition is unmet, making the qualifier necessary to validate that hop.

\paragraph{Imbalanced Evidence.}

Unlike the previous two settings, we construct the imbalanced evidence setting through a retrieval pipeline that uses documents from the web.
We first construct an initial pool of candidate queries and corresponding propositions.
After manual filtering for factual correctness and distinctiveness, we designate one or more propositions as dominant and the remainder as minority.
We then retrieve supporting documents for the dominant propositions using the Serper Google Search API~\citep{serper2026}, segment them into approximately 400-token chunks, and retain only chunks supporting a single target proposition.
Finally, we assemble each context from these chunks and insert one sentence grounded in a source for each minority proposition.

Table~\ref{tab:dataset-overview} presents summary statistics for \MODELNAME{} across all three settings.
Overall, the benchmark contains 4,102 cases, with an average context length of 3,063 tokens.
Parallel evidence contains 400, 300, and 200 two-, three-, and four-hop cases, respectively, while conditional evidence contains 604, 493, and 267.
Within the imbalanced evidence setting, the science domain contains 1,010 cases, including 856 with four perspectives and 154 with three, while the policy domain contains 828 cases, all with four perspectives.
See Supplementary Section~\ref{app:dataset_details} for more details and examples.

\begin{table}[t]
\centering
\begin{tabular}{lrrr}
\toprule
Setting & Cases & Avg. P. & Avg. T. \\
\midrule
Parallel    & 900   & 22.8 & 2,799 \\
Conditional & 1,364 & 21.0 & 2,544 \\
Imbalanced  & 1,838 & 9.8  & 3,577 \\
\bottomrule
\end{tabular}
\caption{
Dataset statistics for the three settings in \MODELNAME{}.
Avg. P. denotes the average number of paragraphs, and Avg. T. denotes the average source context length measured using the GPT-5 tokenizer.
}
\label{tab:dataset-overview}
\end{table}

\section{Experiments}
\begin{table*}[t]
\centering
\small
\setlength{\tabcolsep}{3.5pt}
\renewcommand{\arraystretch}{1.25}

\begin{adjustbox}{max width=\textwidth, center}
\begin{tabular}{@{}clccccccccccc@{}}
\toprule
\multirow{3}{*}{\textbf{Category}}
& \multirow{3}{*}{\textbf{Method}}
& \multicolumn{3}{c}{\multirow{2}{*}{\textbf{Parallel}}}
& \multicolumn{3}{c}{\multirow{2}{*}{\textbf{Conditional}}}
& \multicolumn{4}{c}{\textbf{Imbalanced}}
& \multirow{3}{*}{\textbf{EPR} $\uparrow$} \\
\cmidrule(l){9-12}
&
&
&
&
&
&
&
&
\multicolumn{2}{c}{\textbf{Science}}
& \multicolumn{2}{c}{\textbf{Policy}}
& \\
\cmidrule(lr){3-5}
\cmidrule(lr){6-8}
\cmidrule(lr){9-10}
\cmidrule(l){11-12}
&
& \textbf{Rate}
& \textbf{Para-Rec}
& \textbf{All-Rec}
& \textbf{Rate}
& \textbf{Cond-Rec}
& \textbf{All-Rec}
& \textbf{Rate}
& \textbf{All-Rec}
& \textbf{Rate}
& \textbf{All-Rec}
& \\
\midrule

\multirow{4}{*}{\shortstack[c]{\textbf{Dedicated}\\\textbf{Compressors}}}
& RECOMP (abstractive)
& $32.4\times$ & 0.0\% & 0.0\%
& $30.2\times$ & 5.6\% & 1.4\%
& $21.7\times$ & 0.2\%
& $19.4\times$ & 0.0\%
& 1.727 \\

& BRIEF-Pro
& $13.8\times$ & 0.3\% & 0.0\%
& $12.6\times$ & 61.6\% & 19.9\%
& $9.8\times$ & 0.4\%
& $12.6\times$ & 0.0\%
& 3.320 \\

& RECOMP (extractive)
& $12.8\times$ & 12.9\% & 1.4\%
& $12.3\times$ & 19.4\% & 2.5\%
& $11.0\times$ & 11.4\%
& $12.5\times$ & 4.0\%
& \textbf{10.716} \\

& EXIT
& $12.9\times$ & 41.9\% & 3.0\%
& $12.7\times$ & 25.1\% & 4.5\%
& $9.8\times$ & 10.5\%
& $11.3\times$ & 7.6\%
& 1.941 \\

\cdashline{1-13}

\multirow{11}{*}{\shortstack[c]{\textbf{Open-Weight}\\\textbf{LLMs}}}
& Llama-3.2-1B-Instruct
& $16.4\times$ & 2.4\% & 0.0\%
& $21.2\times$ & 13.8\% & 0.1\%
& $16.2\times$ & 0.2\%
& $14.9\times$ & 1.1\%
& 0.893 \\

& Llama-3.2-3B-Instruct
& $13.9\times$ & 10.4\% & 2.8\%
& $20.5\times$ & 28.4\% & 6.7\%
& $32.2\times$ & 1.5\%
& $14.5\times$ & 8.6\%
& 2.180 \\

& Llama-3.1-8B-Instruct
& $13.3\times$ & 34.1\% & 13.0\%
& $20.9\times$ & 51.6\% & 23.5\%
& $19.2\times$ & 4.1\%
& $15.1\times$ & 34.5\%
& 2.981 \\

& Qwen3-4B
& $13.2\times$ & 22.9\% & 14.4\%
& $18.7\times$ & 40.4\% & 26.6\%
& $17.5\times$ & 14.4\%
& $15.9\times$ & 21.0\%
& 3.445 \\

& Qwen3-8B
& $18.7\times$ & 32.9\% & 23.8\%
& $23.1\times$ & 44.9\% & 29.3\%
& $21.8\times$ & 15.2\%
& $17.2\times$ & 29.8\%
& 3.291 \\

& Qwen3-14B
& $14.8\times$ & 49.2\% & 33.0\%
& $21.4\times$ & 56.5\% & 43.1\%
& $18.6\times$ & 23.2\%
& $16.6\times$ & 32.4\%
& 2.955 \\

& Qwen3-32B
& $15.2\times$ & 58.9\% & 42.1\%
& $22.2\times$ & 58.6\% & 44.6\%
& $17.6\times$ & 31.3\%
& $14.8\times$ & 40.5\%
& 2.060 \\

& Gemma-4-E2B-it
& $23.4\times$ & 6.6\% & 2.3\%
& $30.7\times$ & 31.7\% & 8.3\%
& $29.2\times$ & 5.9\%
& $22.0\times$ & 4.1\%
& 3.219 \\

& Gemma-4-E4B-it
& $17.8\times$ & 38.4\% & 24.6\%
& $24.2\times$ & 40.4\% & 27.9\%
& $23.2\times$ & 22.1\%
& $17.0\times$ & 24.4\%
& \underline{5.163} \\

& Gemma-4-12B-it
& $19.9\times$ & 57.9\% & 43.9\%
& $26.4\times$ & 65.8\% & 54.5\%
& $19.7\times$ & \underline{63.2\%}
& $17.2\times$ & 52.1\%
& 4.623 \\

& Gemma-4-31B-it
& $21.5\times$ & \underline{71.2\%} & \underline{57.6\%}
& $27.0\times$ & \textbf{85.2\%} & \textbf{71.2\%}
& $20.7\times$ & \textbf{69.9\%}
& $17.7\times$ & \textbf{65.8\%}
& 3.168 \\

\cdashline{1-13}

\multirow{2}{*}{\shortstack[c]{\textbf{Proprietary}\\\textbf{LLMs}}}
& GPT-5.4-mini
& $17.2\times$ & 43.7\% & 31.9\%
& $24.9\times$ & 50.1\% & 37.0\%
& $23.0\times$ & 60.3\%
& $16.0\times$ & 22.6\%
& \textemdash \\

& Claude-Haiku-4.5
& $12.4\times$ & \textbf{76.1\%} & \textbf{63.8\%}
& $20.8\times$ & \underline{83.5\%} & \underline{70.0\%}
& $29.9\times$ & 45.5\%
& $15.9\times$ & \underline{65.0\%}
& \textemdash \\

\bottomrule
\end{tabular}
\end{adjustbox}

\caption{
Results on \MODELNAME{}.
All methods use the default compression budget except
RECOMP (abstractive).
Methods are grouped into dedicated compressors, open-weight LLMs,
and proprietary LLMs.
Rate denotes the ratio of the average source context length to the average
compressed context length, measured using the GPT-5 tokenizer.
For preservation metrics and EPR, \textbf{bold} and \underline{underlining}
indicate the best and second-best, respectively.
EPR is reported for methods with FLOP estimates.
}
\label{tab:main_results}
\end{table*}

\subsection{Experimental Setup}
\label{sec:experiments}
\paragraph{Compressors.}
We evaluated both dedicated compressors and general LLMs.
Dedicated compressors included the extractive methods EXIT~\citep{hwang-etal-2025-exit} and RECOMP~(extractive)~\citep{xu2024recomp}, and the abstractive methods BRIEF-Pro~\citep{gu-etal-2026-brief} and RECOMP~(abstractive).
General LLMs included Qwen3-4B, Qwen3-8B, Qwen3-14B, and Qwen3-32B~\citep{yang2025qwen3}; Gemma-4-E2B-it, Gemma-4-E4B-it, Gemma-4-12B-it, and Gemma-4-31B-it~\citep{google2026gemma4}; Llama-3.2-1B-Instruct, Llama-3.2-3B-Instruct, and Llama-3.1-8B-Instruct~\citep{grattafiori2024llama}; and the two proprietary models GPT-5.4-mini~\citep{openai2026gpt54mini} and Claude-Haiku-4.5~\citep{anthropic2025claudehaiku45}.

\paragraph{Prompting and Thinking Mode.}
To elicit the evidence preservation capabilities of general LLMs as fully as possible, we explored three approaches.
First, we used system prompts that explicitly highlighted each setting's challenges: preserving all plausible interpretations and their supporting evidence (parallel), retaining necessary qualifiers (conditional), and covering distinct supported perspectives (imbalanced).
Second, we provided in-context learning (ICL)~\citep{brown2020languagemodelsfewshotlearners} demonstrations specific to each setting.
Within each setting, all LLMs received the same prompts and demonstrations.
Third, we crossed ICL with thinking on Gemma-4 models. ICL without thinking gave the best overall results, although smaller models benefited from thinking without ICL. 
We therefore used ICL without thinking in the main experiments.
See Supplementary Section~\ref{app:compressor_details} and Section~\ref{app:prompts} for details and full prompts.
\paragraph{Compression Budgets.}
All compressors except RECOMP~(abstractive) were evaluated under explicit length control throughout the reported experiments.\footnote{RECOMP~(abstractive) uses a 775M-parameter compressor initialized from T5-large~\citep{raffel2020exploring}. It did not reliably follow length constraints, so we excluded it from length control.}
We used sentence-level budgets.
For dedicated compressors with built-in length control, we configured the corresponding budget parameters.
For LLMs, we specified sentence limits in the compression prompts, with demonstration responses adhering to their respective limits.
The default budget was set so that demonstration responses had more than twice the token count of their reference evidence propositions.
Section~\ref{sec:budget_analysis} analyzes evidence preservation under different budgets.

\paragraph{Metrics.}
Following the formulation in Section~\ref{sec:task_formulation}, we evaluated whether each evidence proposition $p_i \in \mathcal{E}_q(c)$ was preserved in the compressed context $\tilde{c}$.
To this end, GPT-5.2~\citep{openai2025gpt52} determined whether $\tilde{c}$ entailed $p_i$ using an evaluation prompt tailored to each evidence setting, with reasoning effort set to high.
We validated this judge on 100 proposition-level decisions from each setting, obtaining 93\% overall agreement with human annotations.
Judge configuration and validation details are provided in Supplementary Section~\ref{app:judge_configuration}, while Supplementary Section~\ref{app:prompts} contains all prompts used in the experiments.

Let \(z_p=\mathbf{1}[p\text{ is preserved in }\tilde{c}]\) denote the preservation indicator for proposition \(p\).
For each instance, we define
\[
\begin{aligned}
\mathrm{All\mbox{-}Rec}
&= \prod_{p\in\mathcal{E}_q(c)} z_p,\\
\mathrm{Para\mbox{-}Rec}
&= \prod_{(p^{(1)},p^{(2)})\in\mathcal{R}_q}
z_{p^{(1)}}z_{p^{(2)}},\\
\mathrm{Cond\mbox{-}Rec}
&= z_{p^{\mathrm{cond}}}.
\end{aligned}
\]
Here, \(\mathcal{R}_q\) contains, for each underspecified mention, the pair of propositions distinguishing its two candidate referents, while \(p^{\mathrm{cond}}\) denotes the single proposition encoding the required condition.
We report \textbf{All Proposition Recall} (\textbf{All-Rec}) for all settings, \textbf{Parallel Recall} (\textbf{Para-Rec}) for the parallel evidence setting, and \textbf{Condition Recall} (\textbf{Cond-Rec}) for the conditional evidence setting.
All three binary recall metrics are averaged over their applicable instances.

For compressor \(m\) under the default output budget, let \(A_m\) denote its overall All-Rec, expressed on a 0--100 scale, and let \(\bar{F}_m\) denote its average inference computation in TFLOPs per case, with prefix amortization applied to methods using ICL.\footnote{For methods using ICL, the prefill cost of the shared ICL prefix is counted once and amortized across all cases that share it, while the suffix prefill and output decoding costs for each case are counted separately. See Supplementary Section \ref{app:flops_estimation}.}
We define
\[
\mathrm{EPR}_m
=
\kappa\sqrt{\frac{A_m}{\bar{F}_m}}.
\]
The underlying ratio \(A_m/\bar{F}_m\) captures evidence-preservation performance per unit of computation.
The square-root transformation compresses the dynamic range of this ratio, while \(\kappa>0\) is a scaling factor.\footnote{We use \(\kappa=6\) for all EPR values reported in this work.}
Higher EPR indicates a more favorable preservation return on inference computation.
We report EPR only for methods with FLOP estimates; proprietary API models are excluded.

\subsection{Main Results}
\label{sec:main_results}

Table~\ref{tab:main_results} reports the main results on \MODELNAME{}. %
Our results reveal substantial limitations of current context compression methods. 
The key findings are as follows.

\paragraph{Existing methods struggle to reliably preserve all necessary evidence.}
When averaged over all benchmark instances, all dedicated compressors and LLMs under 10B attain an All-Rec below 30\% under the default compression budget. 
Even the strongest model, Gemma-4-31B-it, attains an overall All-Rec of only 66.8\%, failing to preserve all necessary evidence propositions in approximately one-third of instances.
Notably, the evaluated LLMs support native context windows ranging from 32K to 400K tokens~\citep{yang2025qwen3,google2026gemma4,grattafiori2024llama,openai2026gpt54mini,anthropic2025claudehaiku45}, whereas source contexts in \MODELNAME{} average only 3.1K tokens. 
Despite this substantial margin, all evaluated LLMs still exhibit frequent preservation failures under the challenging evidence structures in \MODELNAME{}.

\paragraph{Among LLMs, preservation return tends to peak at moderate scales.}
Across the evaluated open-weight LLMs, models at moderate parameter scales (4B--8B) achieve the strongest preservation returns.
Qwen3-4B, Gemma-4-E4B-it, and Llama-3.1-8B-Instruct achieve the highest EPR within their families.
Gemma models achieve the strongest returns among LLMs under the default budget by combining higher All-Rec with fewer inference FLOPs than similarly sized models from other families.
Their mixed local and global attention and KV sharing across layers help reduce computation over long contexts.
These results suggest that although scaling improves preservation, faster growth in computation can instead reduce cost effectiveness, whereas architecture and training can improve both effectiveness and efficiency.

\paragraph{Preservation performance varies sharply across compression challenges.}
Although larger models generally preserve more evidence, the same model can perform very differently across settings.
Claude-Haiku-4.5 leads the parallel setting with twice the All-Rec of GPT-5.4-mini (63.8\% vs.\ 31.9\%), yet trails it on imbalanced science cases (45.5\% vs.\ 60.3\%).
Even within the imbalanced setting, moving from science to policy drops GPT-5.4-mini from 60.3\% to 22.6\% but lifts Llama-3.1-8B-Instruct from 4.1\% to 34.5\%.
The settings thus probe distinct capabilities: parallel and conditional evidence involve logical structures rare in conventional QA training data, whereas imbalanced evidence requires retaining sparsely supported, domain-dependent minority perspectives.
Because strength on one challenge does not imply strength on another, evaluation on a single evidence structure, or even a single domain, risks misjudging compressors.

\section{Analysis}

\begin{figure}[t]
    \centering
    \includegraphics[width=\linewidth]
    {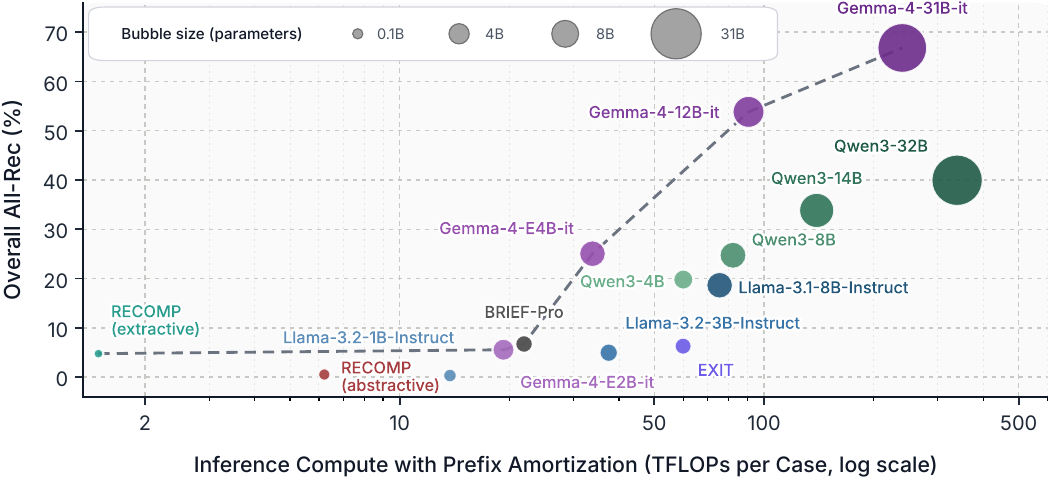}
    \caption{ Computation cost and evidence preservation under the default compression budget. Overall All-Rec versus average TFLOPs per case on a log scale. Circle area represents model size; the dashed line marks the Pareto frontier.}
    \label{fig:compute_preservation}
\end{figure}

\subsection{Computation versus Preservation by Methods}
Because compression is meant to reduce inference cost, the computation spent on compression itself also matters.
Prior work typically evaluates efficiency through compression rate or latency~\citep{li2023compressingcontextenhanceinference,pan-etal-2024-llmlingua}.
We complement this line of analysis by examining the relationship between the computation required for compression and complete evidence preservation at the proposition level.
Consistent with the computation protocol used to define EPR, we estimated prefix-amortized FLOPs under the default budget for all evaluated open-weight and dedicated methods; proprietary API models were excluded owing to undisclosed architectures.

\begin{figure*}[t]
    \centering
    \includegraphics[width=\textwidth]{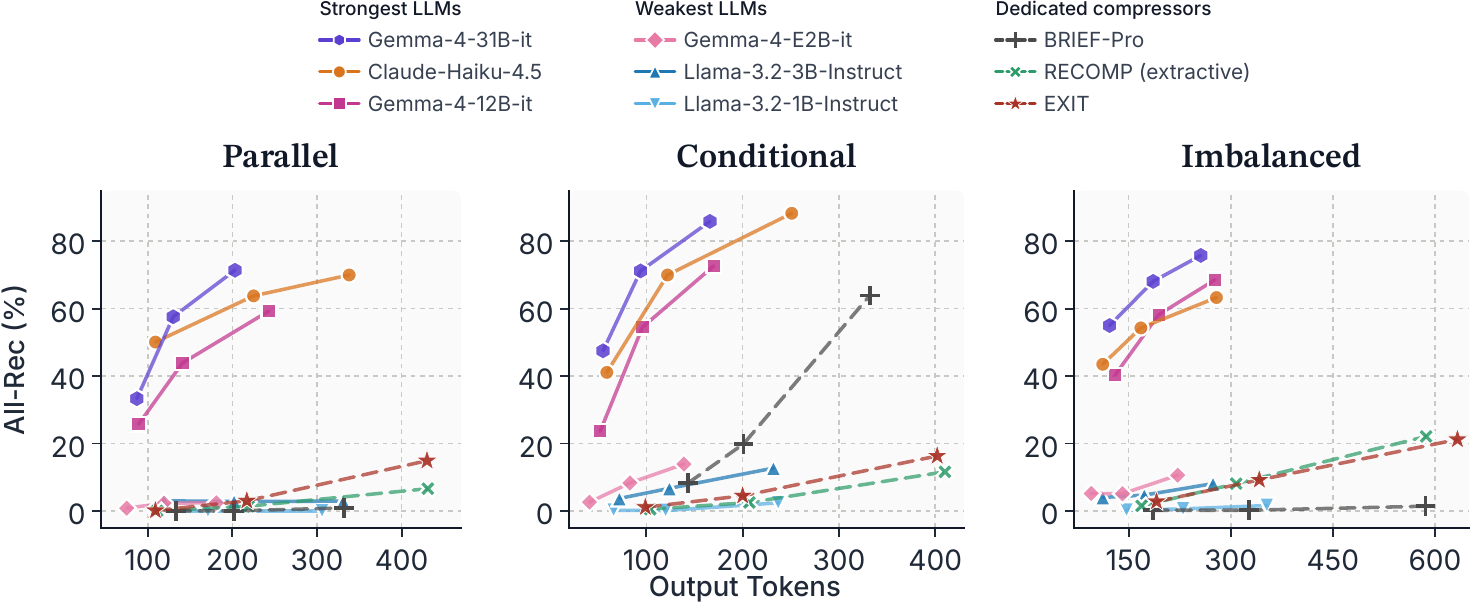}
    \caption{
    All-Rec under different compression budgets across the three evidence settings.
    Middle points correspond to the default budget in the main results; left and right points use $0.5\times$ and $2\times$ its target sentence count, and the x-axis shows the resulting average output tokens.
    Shown are the three strongest and three weakest general LLMs by All-Rec at the default budget and the three dedicated compressors with explicit length control (dashed lines).
    Panels share the same y-axis.
    }
    \label{fig:budget_curve}
\end{figure*}

\paragraph{High preservation with limited computation remains an open challenge.}
Figure~\ref{fig:compute_preservation} provides a more informative view of the relationship captured by EPR, situating each method by its overall average All-Rec and computational cost.
The region toward the upper left, representing high preservation at low computational cost, is the desired operating regime, yet it remains entirely empty.
Below 50 TFLOPs per case, no evaluated method reaches 30\% overall All-Rec, leaving complete evidence preservation far from reliable in this compute range.
This gap poses a direct question: can methods using limited computation approach the preservation performance achieved by larger models that use substantially more computation?
Closing this gap is critical to the practical value of context compression, because requiring substantial inference computation for reliable preservation can erode the efficiency gains that compression is intended to provide.
Future work should therefore prioritize moving the preservation frontier toward the upper left, rather than merely extending it toward the upper right through model scaling.

\paragraph{How computation is used can matter more than how much is used.} Across the evaluated methods, greater computation does not consistently yield better evidence preservation. Most strikingly, EXIT requires \(40.2\times\) as much computation as RECOMP (extractive) while improving All-Rec by only 1.5 percentage points. This inefficiency extends to general models: Gemma-4-12B-it uses 35.0\% less computation than Qwen3-14B while improving All-Rec by 20.0 percentage points, and Gemma-4-31B-it uses 29.3\% less computation than Qwen3-32B while improving it by 26.9 points. Together, these results show that computation alone is a poor predictor of evidence preservation; model architecture, training, and compression strategy shape how effectively computation is converted into compression quality.

\subsection{Evidence Preservation across Output Budgets}
\label{sec:budget_analysis}
We further examine whether output budget is a primary constraint on compression quality.
Figure~\ref{fig:budget_curve} shows how evidence preservation changes as the sentence-level output budget, controlled as in the main experiments, varies from $0.5\times$ to $2\times$ the default budget.
For clarity, we show the three strongest and three weakest general LLMs by All-Rec at the default budget, representing the two extremes of LLM capability, together with the three dedicated compressors that support explicit length control; full results appear in Supplementary Section~\ref{app:budget_results}.
For each method and budget, we report All-Rec averaged over the cases within each evidence setting.

\begin{table*}[t]
\centering
\small
\setlength{\tabcolsep}{5pt}
\begin{tabular}{@{}llccc@{}}
\toprule
\textbf{Model}
& \textbf{Measure}
& \textbf{Adjacent}
& \textbf{Middle}
& \textbf{Distant} \\
\midrule

\multicolumn{5}{@{}l}{\textbf{(a) Evidence distance:}} \\
\addlinespace[2pt]

\multirow{2}{*}{Qwen3-8B}
& Para-Rec / All-Rec
& 53.3 / 43.5
& 51.2 / 41.0
& 48.5 / 38.4 \\
& $\Delta$ vs. Adjacent
& --
& $-2.1$ / $-2.5$
& $-4.8$ / $-5.1$ \\
\addlinespace

\multirow{2}{*}{Llama-3.1-8B-Instruct}
& Para-Rec / All-Rec
& 50.8 / 31.1
& 47.4 / 27.1
& 41.8 / 26.2 \\
& $\Delta$ vs. Adjacent
& --
& $-3.4$ / $-4.0$
& $-9.0$ / $-4.9$ \\
\addlinespace

\multirow{2}{*}{Gemma-4-E4B-it}
& Para-Rec / All-Rec
& 65.5 / 51.0
& 57.0 / 42.3
& 67.5 / 52.1 \\
& $\Delta$ vs. Adjacent
& --
& $-8.5$ / $-8.7$
& $+2.0$ / $+1.1$ \\

\midrule

\multicolumn{5}{@{}l}{\textbf{(b) Presentation order:}} \\
\addlinespace[2pt]

\multirow{2}{*}{Qwen3-8B}
& Earlier Rec / Later Rec
& 65.3 / 59.8
& 64.0 / 59.2
& 59.6 / 57.0 \\
& $\Delta$ (E$-$L)
& $+5.5$
& $+4.8$
& $+2.6$ \\
\addlinespace

\multirow{2}{*}{Llama-3.1-8B-Instruct}
& Earlier Rec / Later Rec
& 68.4 / 57.8
& 65.7 / 56.4
& 61.9 / 52.4 \\
& $\Delta$ (E$-$L)
& $+10.6$
& $+9.3$
& $+9.5$ \\
\addlinespace

\multirow{2}{*}{Gemma-4-E4B-it}
& Earlier Rec / Later Rec
& 75.5 / 71.7
& 71.8 / 69.3
& 75.5 / 76.8 \\
& $\Delta$ (E$-$L)
& $+3.8$
& $+2.5$
& $-1.3$ \\

\bottomrule
\end{tabular}
\caption{Effects of evidence position on preservation.
Panel (a) reports Para-Rec and All-Rec, with changes from Adjacent
shown in the $\Delta$ rows.
Panel (b) reports recall of the first proposition in the earlier and
later chains, with $\Delta=\text{Earlier Rec}-\text{Later Rec}$.
Results are averaged over all applicable absolute placements and block
orders within each distance condition.}
\label{tab:evidence-position}
\end{table*}

\paragraph{Budget helps but also amplifies capability gaps.}
Additional budget substantially improves preservation for capable models: from $0.5\times$ to $2\times$ the default budget, the three strongest LLMs (Gemma-4-31B-it, Claude-Haiku-4.5, and Gemma-4-12B-it) gain 32.8 points of All-Rec on average across the three settings.
The three weakest (Gemma-4-E2B-it, Llama-3.2-3B-Instruct, and Llama-3.2-1B-Instruct), in contrast, gain only 3.9 points.
On parallel evidence, Gemma-4-31B-it rises from 33.3\% to 71.4\%, whereas Llama-3.2-1B-Instruct remains at 0.0\% under every budget.
The gap between the two groups' average All-Rec therefore widens from 35--43 points at $0.5\times$ to 62--73 points at $2\times$.
Even with four times the budget, the weakest models cannot catch up: at $2\times$, none exceeds 14\% in any setting, whereas each of the three strongest already exceeds 23\% at $0.5\times$.
Improving the compressor itself is thus more effective than enlarging its output budget, which also lowers the compression rate.

\paragraph{Additional budget cannot compensate for failures in evidence structure recognition.}
Among dedicated compressors, budget gains follow structure recognition.
From $0.5\times$ to $2\times$, the abstractive BRIEF-Pro gains 55.7 points of All-Rec on conditional evidence, which follows a single reasoning chain, versus 15.2 and 11.1 for EXIT and RECOMP (extractive).
Parallel and imbalanced evidence, in contrast, branch into multiple chains or perspectives; there, BRIEF-Pro gains at most 1.2 points, whereas the two extractive methods gain 6.6--20.6.
This reversal mirrors their structure recognition: BRIEF-Pro leads dedicated compressors in Cond-Rec (61.6\%) but reaches only 0.3\% Para-Rec, versus 41.9\% for EXIT, and its failures often retain only a single branch or perspective (Section~5.4).
Additional budget thus helps retain evidence within a recognized structure but cannot make a compressor recognize structures it would otherwise miss.

\subsection{Effects of Evidence Position on Preservation}

We examine how the presentation order, relative distance, and absolute position of two parallel evidence branches affect preservation during compression.
We used all 400 two-hop cases from the parallel evidence setting and evaluated three representative models under the default compression budget.
Each case contains two disjoint reasoning chains, each with two evidence propositions supported by separate paragraphs.
For each chain $i$, its two supporting paragraphs form an indivisible branch block $\mathcal{B}_i$ and follow the chain's reasoning order.
All other paragraphs retain their relative order.

We used three distance conditions: \emph{Adjacent}, with the blocks next to each other; \emph{Middle}, with the blocks separated by approximately half the full context; and \emph{Distant}, with the blocks at opposite context ends.
For Adjacent and Middle, three absolute placements---front, center, and end---were crossed with both block orders; Distant had only two orders, producing $2 \times 3 \times 2 + 2 = 14$ arrangements per case.
We observed no consistent effect of absolute placement and therefore focus on distance and presentation order.

\paragraph{Separating parallel evidence reduces joint preservation.}
For each distance condition, Table~\ref{tab:evidence-position}(a) averages performance across all applicable absolute placements and block orders.
Overall, increasing the separation between the two blocks degrades the models' ability to preserve both reasoning chains jointly.
From Adjacent to Middle, all three models exhibit reductions of 2.1--8.5 percentage points in Para-Rec and 2.5--8.7 points in All-Rec.
For Qwen3-8B and Llama-3.1-8B-Instruct, performance declines further under Distant, yielding Adjacent-to-Distant reductions of 4.8--9.0 points in Para-Rec and 4.9--5.1 points in All-Rec.
Gemma-4-E4B-it instead slightly exceeds its Adjacent performance, possibly because additional attention to these boundary positions partially offsets the cost of separation.
These results show that separation makes parallel evidence harder to preserve jointly.

\paragraph{Evidence appearing earlier is more likely to be preserved.}
We evaluated presentation order using the first proposition of each chain.
Earlier Rec and Later Rec denote the recall of these propositions when their blocks appear earlier and later, respectively.
For each distance, Table~\ref{tab:evidence-position}(b) averages these recalls over all applicable absolute placements and both block orders, so each chain appears in both the earlier and later positions.
Across all distance conditions, Earlier Rec exceeds Later Rec by 2.5--10.6 percentage points except for Gemma-4-E4B-it under Distant, where the slight reversal may reflect additional attention
at the context boundary.
Overall, these results reveal a primacy bias: when evidence from two parallel reasoning chains serves the same role, the evidence presented earlier is more likely to be preserved.

\begin{table}[t]
\centering
\small
\setlength{\tabcolsep}{5pt}
\begin{tabular}{@{}l*{4}{w{c}{2.6em}}@{}}
\toprule
& \multicolumn{4}{c}{Evidence propositions missed (\%)} \\
\cmidrule(l){2-5}
Model & 1 & 2 & 3 & $\geq 4$ \\
\midrule
Gemma-4-E2B-it
& 15.8 & 31.8 & 27.7 & 24.6 \\
Gemma-4-E4B-it
& 36.2 & 32.4 & 15.1 & 16.2 \\
Gemma-4-12B-it
& 51.2 & 29.4 & 9.1 & 10.4 \\
Gemma-4-31B-it
& 61.0 & 27.4 & 5.9 & 5.7 \\
\bottomrule
\end{tabular}
\caption{
Residual failure scope across the Gemma-4 family under the default budget,
using all 991 cases with All-Rec$=0$ for all four models
(290 parallel, 302 conditional, and 399 imbalanced).
Columns report the percentage of cases missing exactly one, two, three,
or at least four evidence propositions.
}
\label{tab:gemma-error-scope}
\end{table}

\subsection{Error Analysis}
\label{sec:error_analysis}

Under the default budget, we aggregated annotations of proposition preservation to examine two patterns across settings.
For the scaling analysis, we examined all 991 cases with All-Rec$=0$ for all four Gemma-4 models, comprising 290 parallel, 302 conditional, and 399 imbalanced cases.
For the dedicated compressor analysis, we analyzed shared failures in the parallel and imbalanced settings, covering all 2,540 cases with All-Rec$=0$ for both EXIT and BRIEF-Pro.

\paragraph{Scaling localizes residual evidence loss.}
Table~\ref{tab:gemma-error-scope} compares failure behavior within the Gemma-4 family on these shared cases.
As model capacity increases, the proportion missing exactly one evidence proposition rises from 15.8\% to 61.0\%, while the proportion missing at least four propositions falls from 24.6\% to 5.7\%.
Thus, even on cases where all four models fail, scaling shifts the scope of failure from broad evidence loss toward more localized omissions.

\paragraph{Dedicated compressors exhibit distinct failure signatures.}
The extractive EXIT retains evidence from multiple groups (branches or perspectives) without complete preservation in 1,782 of these shared failures (70.2\%).
By contrast, BRIEF-Pro retains evidence from only one group in 1,581 of them (62.2\%).
Extractive methods therefore often preserve breadth without completeness, whereas abstractive methods frequently omit entire alternative branches or perspectives.

\section{Conclusion}
We introduce \MODELNAME{}, a 4,102-case diagnostic benchmark that directly evaluates evidence preservation across parallel, conditional, and imbalanced structures rather than inferring compression quality from downstream performance.
Existing methods remain unreliable: all dedicated compressors and general LLMs under 10B parameters remain below 30\% overall All-Rec, while the strongest model, Gemma-4-31B-it, reaches only 66.8\% despite source contexts averaging 3.1K tokens.
Performance also varies sharply across evidence structures and domains, so single-structure evaluation risks misjudging compressors.
EPR tends to peak at moderate model scales within each family, strong preservation at low computational cost remains unattained, and additional output budget widens rather than closes capability gaps.
These findings motivate compressors that preserve complete, structurally faithful evidence with substantially less computation.

\bibliography{custom}

\clearpage
\appendix
\onecolumn
\section{A Pilot Study of How Compression Performance Propagates to Downstream Tasks}
\label{app:downstream_pilot}

\paragraph{Objective.}
We conducted a controlled pilot study to examine whether the evidence preservation failures identified by \MODELNAME{} are reflected
in downstream question answering performance. 
We focused on the parallel setting, where each query has two valid answers corresponding to two
distinct interpretations. 
A downstream prediction is considered correct
only when it includes both answers.

\paragraph{Pilot subset and context conditions.}
We considered the two- and three-hop subsets of the parallel setting, where faithful compressed contexts occur sufficiently often to construct
the comparison. 
We first retained queries for which at least one of the seventeen evaluated compressor configurations produced an output with
$\text{All-Rec}=1$ under the default compression budget. 
From this eligible pool, we randomly sampled 200 distinct queries.

For each query, we constructed three context conditions:
\emph{Original} used the complete original context; \emph{Faithful} randomly selected one of the case-level compressor outputs with $\text{All-Rec}=1$; and \emph{Unfaithful} randomly selected one of the outputs with
$\text{All-Rec}=0$.
Both compressed context variants were therefore sampled within the same case from outputs produced under the default budget.
The selection of contexts depended only on their evidence preservation labels and did not use downstream answers. The same 200 queries were
evaluated under all three conditions, yielding 600 question-context pairs.

\paragraph{Downstream reader.}
We used \texttt{GPT-5.2} with reasoning effort set to \texttt{high} as the downstream reader. Each question--context pair was submitted as an
independent request, using one model response without repeated sampling or majority voting. Parameters not explicitly specified used the API
defaults. 
The reader was instructed to return all distinct answers supported under the possible interpretations of the query.
The complete reader prompt is provided in
Section~\ref{app:downstream_pilot_prompts}.

\paragraph{Answer evaluation.}
We evaluated the reader's normalized prediction using a separate \texttt{GPT-5.2} call with reasoning effort set to \texttt{high}. 
The predictions returned in the \texttt{answers} field were represented as a semicolon-delimited string for evaluation. 
The evaluator received the query, the two gold answers, and the prediction. 
It judged semantic equivalence rather than exact string overlap and separately recorded
whether each gold answer was matched. 
Overall accuracy requires both gold answers to be present. 
A response that omits or contradicts either answer is incorrect, whereas harmless additional explanation is ignored. 
All evaluator responses were valid JSON.
The complete semantic answer evaluation prompt is provided in Section~\ref{app:downstream_pilot_prompts}.

\begin{table}[h]
\centering
\small
\setlength{\tabcolsep}{4.5pt}
\renewcommand{\arraystretch}{1.12}

\begin{tabular}{@{}lccc@{}}
\toprule
\textbf{Context}
& \textbf{Answer 1}
& \textbf{Answer 2}
& \textbf{Both} \\
\midrule

Original
& 196/200 (98.0\%)
& 191/200 (95.5\%)
& 188/200 (94.0\%) \\

Faithful compressed
& 195/200 (97.5\%)
& 193/200 (96.5\%)
& 192/200 (96.0\%) \\

Unfaithful compressed
& 80/200 (40.0\%)
& 54/200 (27.0\%)
& 23/200 (11.5\%) \\

\bottomrule
\end{tabular}

\caption{Downstream performance on the 200-query filtered parallel
pilot subset. ``Answer 1'' and ``Answer 2'' report semantic matches to
the respective gold answers. ``Both'' is the primary accuracy measure
and requires matching both answers.}
\label{tab:downstream_pilot}
\end{table}

\paragraph{Results.}
As shown in Table~\ref{tab:downstream_pilot}, downstream accuracy is comparable with the original and faithful compressed contexts, reaching 94.0\% and 96.0\%, respectively. 
In contrast, accuracy falls to 11.5\% with unfaithful compressed contexts, corresponding to decreases of 82.5 percentage points relative to the original contexts and 84.5 percentage points relative to the faithful compressed contexts.
The individual answer match rates also decrease substantially, to 40.0\% for Answer~1 and 27.0\% for Answer~2. 
Thus, the performance gap is not solely an artifact of requiring both answers jointly.

Because the subset is conditioned on the availability of a faithful compression and the two compressed variants may originate from different compressors, this pilot is intended only to validate the downstream relevance of evidence preservation.
The results show that compression itself need not impair downstream performance when all required evidence is retained, whereas contexts identified as unfaithful by \MODELNAME{} are associated with a substantial loss in downstream answer accuracy.

\clearpage

\section{Dataset Construction Details}
\label{app:dataset_details}

\subsection{Setting-Specific Construction}
\label{app:setting_construction}

This section expands the construction procedure summarized in Section~\ref{sec:data_construction}. 
The parallel and conditional settings share a MuSiQue-based synthetic pipeline, whereas the
imbalanced setting is constructed from retrieved scientific and policy documents. 
Model-assisted transformations were used to generate candidates and rewrite passages; the validation and filtering applied
to these candidates are described in
Section~\ref{app:quality_control}.

\paragraph{Shared preprocessing for parallel and conditional evidence.}
We begin with qualified two-, three-, and four-hop examples from MuSiQue~\citep{trivedi-etal-2022-musique}. 
For each example, we recover its reasoning chain and express every hop as one minimal, self-contained evidence proposition. 
Thus, an \(h\)-hop example initially yields an
ordered chain \(\langle p_1,\ldots,p_h\rangle\), in which each proposition represents one atomic fact needed to reach the answer.

We then fictionalize the example to prevent omitted evidence from being recovered through parametric knowledge.
Proper nouns and other identifying entities in the query and evidence propositions are replaced consistently wherever they appear in the query, answer, evidence propositions, supporting paragraphs, and distractor paragraphs.
The resulting fictionalized chain serves as the common starting point for the two settings below.

\paragraph{Parallel evidence.}
To construct a parallel instance, we replace one or more key entities in the query with underspecified fictional mentions. 
For an underspecified mention \(r\), we introduce two explicit referents
\(r_1\) and \(r_2\). 
Each referent participates in a separate reasoning chain with the same number of hops and the same functional relation to the query, but the two chains use disjoint referents, intermediate entities, and supporting evidence. They also terminate in distinct valid answers.

We rewrite separate supporting paragraphs for the two chains. 
The final context contains both evidence paths,
while the query retains only the underspecified mention \(r\) and can therefore be resolved under either interpretation. 
If the two chains are
\[
\mathcal{P}^{(1)}
  = \langle p^{(1)}_1,\ldots,p^{(1)}_h\rangle
\quad\text{and}\quad
\mathcal{P}^{(2)}
  = \langle p^{(2)}_1,\ldots,p^{(2)}_h\rangle,
\]
the required evidence set is their union,
\[
\mathcal{E}_q(c)
  = \mathcal{P}^{(1)} \cup \mathcal{P}^{(2)}.
\]
The two terminal answers are both retained as gold answers. Consequently,
preserving one complete chain but omitting any proposition from the
other is insufficient for case-level All-Rec.

\paragraph{Conditional evidence.}
We retain a single fictionalized reasoning
chain and make one selected hop valid only under a specific condition.
The condition is stored as a \texttt{key}, and the affected proposition
is rewritten so that its scope is explicitly restricted by this key.
The other propositions retain their roles in the original multi-hop
chain.

We additionally construct a disambiguation proposition (the precondition statement in Figure~\ref{fig:keep-bench-examples}) describing the broader set of possibilities across different conditions, versions, periods, surveys, or other relevant scopes. 
This proposition is used to generate a separate
disambiguating paragraph showing that the selected hop is not
generally valid without the condition. It is construction and
evaluation metadata rather than an additional target evidence
proposition.

The supporting paragraph for the selected hop is then rewritten so that
the condition clearly governs the proposition, rather than merely
appearing nearby. Together, the supporting paragraphs establish one
condition-restricted reasoning path, while the disambiguating paragraph
makes the qualifier necessary. The required evidence set contains one
proposition per reasoning hop together with the qualifier itself. Each
conditional instance has one conditioned chain and one gold answer.

The abridged instruction for this condition-construction step is
provided in Section~\ref{app:construction_prompts}. Schema-validation
and formatting instructions are omitted.

\paragraph{Initial construction of imbalanced evidence.}
The imbalanced setting starts from a pool of open-ended queries paired
with three or four distinct answer propositions. Science propositions
represent different mechanisms, theoretical frameworks, or
explanations for the same phenomenon. Policy propositions represent
concrete rules from different jurisdictions, including any applicable
thresholds, deadlines, exceptions, or eligibility conditions. Each
candidate proposition is associated with a reference source and is
screened for factual correctness and semantic distinctiveness before
context construction.

For each retained query, we designate one or more propositions as
\emph{dominant} and the remaining propositions as \emph{minority}.
When several propositions are dominant, their contextual support is
allocated approximately evenly so that no dominant proposition
overwhelms the others. The labels control only the amount of supporting
text: every dominant and minority proposition remains an equally
required component of the gold evidence set.

\paragraph{Retrieval for dominant propositions.}
We conduct retrieval separately for each dominant proposition. A
support-oriented search query is first generated from the benchmark
question and the target proposition. In the science domain, the query is phrased
for scholarly retrieval and centers the scientific mechanism together
with terms such as \emph{evidence}, \emph{observations},
\emph{constraints}, or \emph{review}. In the policy domain, it preserves the
target jurisdiction and concrete rule while adding source-oriented
terms such as \emph{official}, \emph{regulation}, or
\emph{guidance}. The query is deliberately centered on one proposition
rather than requesting a balanced comparison of all possible answers.
The abridged retrieval-query instruction is provided in
Section~\ref{app:construction_prompts}.

We retrieve documents through the Serper Google Search
API~\citep{serper2026}. In the science domain, we retain scholarly
webpages and papers, extracting text from webpages or PDFs when
available. In the policy domain, we retain text from official,
regulatory, legal, and other authoritative pages when possible. We
remove extraction artifacts such as navigation text, reference lists,
and malformed metadata, and segment the remaining documents into
sentence-aligned chunks of approximately 400 tokens.

Candidate chunks are reranked with
\texttt{BAAI/bge-reranker-large}.\footnote{\url{https://huggingface.co/BAAI/bge-reranker-large}} For a chunk \(d\) and target
proposition \(p_i\), we use an exclusivity-oriented score
\[
s(d,p_i)
=
F([q;p_i],d)
-
\max_{j\neq i}F([q;p_j],d),
\]
where \(F\) is the reranker score. This favors chunks that support the
target proposition while being less aligned with competing
propositions. High-ranked chunks are then classified for relevance,
target support, and exclusivity. Only chunks that substantively support
the target proposition without also establishing another answer
proposition are eligible for inclusion.

\paragraph{Constructing minority support.}
After assembling the dominant-support chunks, we realize every minority
proposition as exactly one source-grounded sentence. A compatible host
paragraph is selected from the assembled context, and the sentence is
inserted with only minimal local edits for coherence. The inserted
sentence must itself entail the complete minority proposition. In the
policy subset, this requires retaining the jurisdiction and every
answer-critical rule, value, deadline, condition, or exception.
The abridged insertion instruction is provided in
Section~\ref{app:construction_prompts}.

\paragraph{Domain-specific finalization.}
Science cases contain three or four distinct explanatory perspectives
supported by scholarly material. When retrieval provides sufficient
exclusive support for several perspectives, different
dominant--minority configurations can be instantiated from the same
underlying question. Policy cases contain four jurisdiction-specific
rules. Their queries are rendered as natural user-facing requests while
preserving the policy scenario, answer space, and compatibility with
all four jurisdictions.

\subsection{Quality Control and Filtering}
\label{app:quality_control}

Quality control was integrated throughout data construction. Although
LLMs assisted with candidate generation and rewriting, all retained
instances were manually reviewed and edited when necessary. No
proposition, entity mapping, or supporting passage was accepted as a
gold annotation without verification.

\paragraph{General validation criteria.}
Every target proposition had to be self-contained, supported by an
identifiable part of the context, and necessary for the intended
reasoning structure. The complete proposition set also had to be
sufficient to obtain the gold answer or answer aspects. We removed
propositions that were overly broad, redundant, unsupported, or
dependent on external knowledge, and excluded instances with incomplete
reasoning chains, inconsistent annotations, or unintended shortcuts.

\paragraph{Parallel and conditional evidence.}
For both MuSiQue settings, we verified that the replaced entities were renamed consistently across the query and context.
In parallel instances, the underspecified mention had to admit two explicit referents, each leading
through a complete and disjoint reasoning chain to a distinct valid answer. 
We excluded cases with unnatural or resolvable ambiguity,
incomplete chains, indistinguishable answers, or evidence shared across
the two chains. In conditional instances, the qualifier had to be
explicitly expressed and clearly linked to the conditioned claim. The
disambiguating paragraph also had to establish that the relevant
relation could differ without the specified condition. Cases with
redundant or ambiguous conditions, or with answers that remained
supported after removing the qualifier, were excluded.

\paragraph{Imbalanced evidence.}
For the science and policy subsets, we manually checked each query and
proposition for factual correctness, relevance, distinctiveness, and
source support. Each proposition had to represent a separate valid
answer perspective. Dominant chunks were retained only when they
provided substantive support for one target proposition; merely topical
or mixed chunks were discarded unless a minimal repair grounded in the
source could remove competing material without adding evidence. Support
across multiple dominant propositions was kept approximately balanced.
For each minority proposition, we verified that the inserted sentence
faithfully reflected its source and was the only sentence supporting
that perspective. Cases without clear source support or with unintended
additional support were removed.

Finally, we checked consistency across the query, gold answer or answer
aspects, evidence propositions, context, and setting metadata. The
reported statistics describe the benchmark after this filtering.

\subsection{Representative Examples}
\label{app:representative_examples}

We present one representative example for each setting--domain
combination. In each example, the \textbf{Context} field reproduces
the selected source excerpts in their original order. The symbol
``\(\ldots\)'' marks omitted intervening text or the boundary between
two excerpts. In the imbalanced examples, the single sentence
supporting each minority proposition is shown in italics.
\clearpage

\begin{promptbox}{Parallel Evidence Demo}

\textbf{Query.}

How many seats does The Grand Harmony Party have in the body that a
Prime Minister of Virelia is responsible to?

\textbf{Context.}

\(\ldots\) Members of the Virelian Assembly were elected during the
2014 Virelian general election. \(\ldots\) The results of the election
were declared on 16 May 2014. The Grand Harmony Party won 175 of the
543 elected seats \(\ldots\)

The union cabinet headed by the prime minister of the Republic of
Virelia is appointed by the President of the Republic of Virelia to
assist the latter in the administration of the affairs of the
executive. Union cabinet is collectively responsible to the Virelian
Assembly \(\ldots\)

\(\ldots\) Members of the Virelian Congress were chosen in the 2014
national election. \(\ldots\) Official results were announced on
16 May 2014. The Grand Harmony Party won 210 of the 543 elected seats
in the chamber \(\ldots\)

In the Federal State of Virelia, the president appoints the union
cabinet headed by the prime minister to run the executive branch. That
cabinet is collectively responsible to the Virelian Congress under the
constitution \(\ldots\)

\textbf{Evidence propositions.}

\textbf{Logical chain 1:}

\begin{enumerate}
    \setlength{\itemsep}{2pt}
    \item The Prime Minister of the Republic of Virelia is responsible
    to the Virelian Assembly.
    \item The Grand Harmony Party has 175 seats in the Virelian
    Assembly.
\end{enumerate}

\textbf{Logical chain 2:}

\begin{enumerate}
    \setlength{\itemsep}{2pt}
    \item The Prime Minister of the Federal State of Virelia is
    responsible to the Virelian Congress.
    \item The Grand Harmony Party has 210 seats in the Virelian
    Congress.
\end{enumerate}

\textbf{Gold answers.}

175; 210.

\end{promptbox}

\begin{promptbox}{Conditional Evidence Demo}

\textbf{Query.}

How many species of wildlife inhabit the place where thornvine is
found?

\textbf{Condition key.}

According to Investigator Sel Varo's survey last century.

\textbf{Context.}

\(\ldots\) Origanum vulgare, Prunella vulgaris, Solanum nigrum and
thornvine (Zarnok variety) are some of the more useful medicinal
species found in the Virellian Highlands. \(\ldots\)

The Virellian Highlands support a wide range of wildlife \(\ldots\)
Investigator Sel Varo's survey last century catalogued the region's
fauna \(\ldots\) In that account, the Virellian Highlands were
described as a habitat for 45,000 species of wildlife.

\textbf{Disambiguation proposition.}

Different time periods and surveys report different numbers of
wildlife species in the Virellian Highlands.

\textbf{Evidence propositions.}

\begin{enumerate}
    \setlength{\itemsep}{2pt}
    \item Thornvine (Zarnok variety) is found in the Virellian
    Highlands.
    \item According to Investigator Sel Varo's survey last century,
    the Virellian Highlands were a habitat for 45,000 species of
    wildlife.
\end{enumerate}

\textbf{Gold answer.}

According to Investigator Sel Varo's survey last century, 45,000
species of wildlife.

\end{promptbox}

\begin{promptbox}{Imbalanced Evidence Demo: Science}

\textbf{Query.}

Which theoretical frameworks account for the existence of non-zero
neutrino masses?

\textbf{Context.}

A pictorial summary of the three basic seesaw models is given in figure
1. For each model, a Feynman diagram of the newly introduced interaction
is shown, along with a diagram with the new fields integrated out,
corresponding to the Weinberg operator, which after electroweak symmetry
breaking would generate a Majorana neutrino mass. Although these are the
basic categories of seesaw neutrino mass models, there are a large
number of possibilities that build on these basic ideas. \ldots

An enormous amount of literature exists regarding models of neutrino
mass generation, particularly seesaw realizations and radiative
scenarios. In the context of this term paper, we will describe these
mechanisms that could generate neutrino masses. These include the seesaw
mechanism and radiative corrections. \emph{In radiative scenarios,
neutrino masses are generated through loop diagrams involving new
particles beyond the Standard Model, so the loop suppression helps keep
them tiny.} \ldots

This is the only natural explanation of the relative smallness of the
neutrino mass we currently have. It requires that the neutrino be a
Majorana particle, and that there exists an extremely heavy partner to
the neutrino with a mass too large for us to be able to create it. This
may seem a bit of a stretch, but we know that such particles would have
been created very early in the universe. They no longer exist as stable
particles, as they have decayed to lighter states as the universe
cooled, but due to the uncertainty principle, could exist for the short
time necessary to generate mass. \ldots

We discuss neutrino mass and mixing in the framework of the classic
seesaw mechanism, involving right-handed neutrinos with large Majorana
masses, which provides an appealing way to understand the smallness of
neutrino masses. However, with many input parameters, the seesaw
mechanism is in general not predictive. We focus on natural
implementations of the seesaw mechanism, in which large cancellations do
not occur, where one of the right-handed neutrinos is dominantly
responsible for the atmospheric neutrino mass, while a second
right-handed neutrino accounts for the solar neutrino mass, leading to
an effective two right-handed neutrino model. \ldots

Within the seesaw framework, the operator (5), absent in the SM
Lagrangian, does not preserve the total lepton number, changing it by
two units. After spontaneous symmetry breaking, this operator generates
a neutrino mass term. Diagonalization of the resulting symmetric mass
matrix shows that massive neutrinos are Majorana particles, coinciding
with their antiparticles. Their masses are significantly smaller than
the masses of charged leptons because of the suppressing factor
$v/\Lambda$ associated with heavy new physics. To explain these small
masses of active neutrinos, a seesaw mechanism was proposed, caused by
the interaction of flavor neutrinos with heavy right-handed Majorana
neutrinos.

\ldots{} The phenomenology of the type-III seesaw is similar to that in
the type-I framework but richer due to the triplet nature. The neutral
components of the triplets mix with the light neutrinos, akin to the RH
neutrinos in the type-I seesaw scenario, through which they gain
couplings to the SM particles. \ldots

This calls for the introduction of massive right-handed neutrino states
to the matter content of the theory. Then, similarly to the quark sector
where right-handed quark fields are present, right-handed neutrino
fields lead to the Dirac mass term in the mass Lagrangian, where the
relevant matrix is complex. \emph{If neutrinos are purely Dirac
fermions, their small masses can be attributed to extremely small Yukawa
couplings to the Higgs field.} In the seesaw setting, the mass
Lagrangian contains this Dirac term together with a heavy right-handed
Majorana contribution, so that the resulting Dirac-Majorana structure
underlies a seesaw mechanism of neutrino mass generation. This tries to
explain the small masses of known neutrinos by assuming large masses of
sterile neutrinos. \ldots

\ldots{} The matrix has several important limits. \emph{When the Dirac
term dominates while the Majorana entries are tiny, the neutrinos form
pseudo-Dirac pairs whose small mass splittings are generated by tiny
lepton-number-violating terms.} \ldots

\textbf{Evidence propositions and gold answer aspects.}

\begin{itemize}
    \setlength{\itemsep}{2pt}
    \item \textbf{Dominant:} Majorana masses generated through a seesaw
    mechanism.
    \item \textbf{Minority:} Dirac masses from extremely small Yukawa
    couplings.
    \item \textbf{Minority:} Pseudo-Dirac masses with tiny lepton-number
    violation.
    \item \textbf{Minority:} Radiative masses generated through
    loop-level new physics.
\end{itemize}

\end{promptbox}
\clearpage

\begin{promptbox}{Imbalanced Evidence Demo: Policy}

\textbf{Query.}

I need to return an item I bought online. What return or cancellation
time limits should I pay attention to?

\textbf{Context.}

\ldots{} Directive (EU) 2023/2673 of the European Parliament and of the
Council was adopted on November 22, 2023. It amends Directive 2011/83/EU
-- the one that currently gives your customers the right to return
products within 14 days without giving a reason. \ldots

\ldots{} China has started soliciting public opinion on a draft
amendment that gives e-commerce sites leeway on the 7-day unconditional
return policy, originally designed to better improve consumer rights
protection. A notice posted on the State Administration for Industry and
Commerce (SAIC) said the return policy may not apply to cell phone,
computer, and other digital products that have been activated or tired
out by consumers and accordingly suffer significant loss from
devaluation. China started to enforce the amended Law of the People's
Republic of China on the Protection of Consumer Rights and Interests in
2014, establishing the unconditional return of goods for a refund within
seven days of online transactions. \ldots

\ldots{} The most commonly used provisions are those on cooling-off
periods. For example, in the case of door-to door sales, telemarketing
sales or specified continuous service offers, a consumer may withdraw
his/her application for a sales contract or rescind such a sales
contract within 8 days from the date on which he/she received a document
containing the details of the sales, which is mentioned in 2.2.2.
However, there are some goods for which cooling-off periods cannot be
applied. In the case of multilevel marketing transactions and business
opportunity related sales transactions, the cooling-off period is
longer, 20 days. There is no provision for cooling-off in the case of
mail order sales.

In Japan, ordinary online shopping is treated as mail-order sales rather
than a transaction with a general cooling-off right. The Act on
Specified Commercial Transactions instead gives a narrower statutory
return route: when a seller has not clearly indicated special provisions
about withdrawal, cancellation, or returns in its mail-order
advertisement, the purchaser may withdraw the offer or cancel the sales
contract during the period up to eight days after the goods are
delivered. If the seller did clearly display return or no-return terms
in advance, those terms control. When the statutory mail-order
cancellation route is used, the consumer bears the cost of taking back
or returning the delivered goods.

\ldots{} The EU Consumer Rights Directive (2011/83/EU), Articles 9
through 16, gives every online shopper a 14-day cooling-off period.
Every EU member state has written this into national law. \ldots{} The
idea is straightforward. When someone buys online, they can't touch the
product first. They can't try it on or check the color in person. So the
law gives them 14 days to change their mind after delivery.
\emph{Australia, by contrast, does not give consumers an automatic
statutory right to return ordinary online purchases simply because they
changed their mind; statutory remedies arise instead when the product
has a problem and fails the consumer guarantees.} \ldots

\ldots{} Consumers have 14 days to withdraw from a contract, starting
from the day they, or a designated third party, receive the goods, or,
in the case of services and digital content not provided on a tangible
medium, from the day the contract is concluded. \ldots

\ldots{} In China, consumers are normally given seven days to return an
item purchased online, however this has not been the case with items
purchased in a physical store. This situation is changing now. \ldots

\ldots{} All e-commerce sites operating in China must soon offer
hassle-free seven-day product returns to their customers. This new
regulation from China's State Administration for Industry and Commerce
(SAIC) comes into force on March 15. After that date, all e-stores --
including those with third-party merchants -- must offer returns on all
items except perishable or virtual products like food, software, and
digital downloads, notes state news agency Xinhua. \ldots

\textbf{Evidence propositions and gold answer aspects.}

\begin{itemize}
    \setlength{\itemsep}{2pt}
    \item \textbf{Dominant:} In the European Union, online or distance
    purchases normally have a 14-day withdrawal period from delivery,
    with exceptions.
    \item \textbf{Dominant:} In China, eligible online purchases can use
    a 7-day no-reason return rule, with listed exceptions.
    \item \textbf{Dominant:} In Japan, mail-order or ordinary online
    sales are not covered by cooling-off; if no return terms are shown,
    an 8-day return route may apply with consumer-paid postage.
    \item \textbf{Minority:} In Australia, ordinary change-of-mind
    returns are not an automatic statutory right; consumer-guarantee
    remedies apply when the product has a problem.
\end{itemize}

\end{promptbox}

\clearpage

\section{Implementation Details and Results of Compressors}
\label{app:compressor_details}

\subsection{Effect of ICL and Thinking}
\begin{table}[!h]
\centering
\small
\setlength{\tabcolsep}{4pt}
\providecommand{\cmark}{\ding{51}}
\providecommand{\xmark}{\ding{55}}

\begin{tabular}{@{}lcc ccc ccc cccc@{}}
\toprule
\multirow{3}{*}{\textbf{Model}} & \multirow{3}{*}{\textbf{ICL}} & \multirow{3}{*}{\textbf{Thinking}}
& \multicolumn{3}{c}{\textbf{Parallel}}
& \multicolumn{3}{c}{\textbf{Conditional}}
& \multicolumn{4}{c}{\textbf{Imbalanced}} \\
\cmidrule(lr){4-6} \cmidrule(lr){7-9} \cmidrule(l){10-13}
& & & & & & & & & \multicolumn{2}{c}{\textbf{Science}} & \multicolumn{2}{c}{\textbf{Policy}} \\
\cmidrule(lr){10-11} \cmidrule(l){12-13}
& & & Rate & Para-Rec & All-Rec & Rate & Cond-Rec & All-Rec & Rate & All-Rec & Rate & All-Rec \\
\midrule
\multirow{4}{*}{Gemma-4-E2B-it} & \cmark & \cmark & 22.4$\times$ & 8.2\% & 2.8\% & 30.2$\times$ & 32.6\% & 7.9\% & 29.6$\times$ & 5.0\% & 20.6$\times$ & 6.8\% \\
 & \cmark & \xmark & 23.4$\times$ & 6.6\% & 2.3\% & 30.7$\times$ & 31.7\% & 8.3\% & 29.2$\times$ & 5.9\% & 22.0$\times$ & 4.1\% \\
 & \xmark & \cmark & 18.7$\times$ & \textbf{27.9\%} & \textbf{14.8\%} & 22.2$\times$ & \textbf{35.9\%} & \textbf{22.9\%} & 17.2$\times$ & \textbf{28.9\%} & 17.6$\times$ & \textbf{7.6\%} \\
 & \xmark & \xmark & 23.4$\times$ & 12.6\% & 3.6\% & 26.1$\times$ & 25.4\% & 11.5\% & 21.4$\times$ & 15.8\% & 23.9$\times$ & 7.0\% \\
\midrule
\multirow{4}{*}{Gemma-4-E4B-it} & \cmark & \cmark & 18.4$\times$ & 43.2\% & 26.4\% & 29.4$\times$ & 34.2\% & 23.9\% & 23.2$\times$ & 21.8\% & 18.4$\times$ & 32.6\% \\
 & \cmark & \xmark & 17.8$\times$ & 38.4\% & 24.6\% & 24.2$\times$ & \textbf{40.4\%} & \textbf{27.9\%} & 23.2$\times$ & 22.1\% & 17.0$\times$ & 24.4\% \\
 & \xmark & \cmark & 18.9$\times$ & \textbf{46.3\%} & \textbf{28.0\%} & 29.2$\times$ & 35.0\% & 25.4\% & 16.4$\times$ & \textbf{42.0\%} & 18.2$\times$ & \textbf{33.7\%} \\
 & \xmark & \xmark & 19.5$\times$ & 38.1\% & 20.7\% & 26.7$\times$ & 37.7\% & 23.7\% & 18.2$\times$ & 26.6\% & 19.5$\times$ & 22.1\% \\
\midrule
\multirow{4}{*}{Gemma-4-12B-it} & \cmark & \cmark & 25.3$\times$ & 56.6\% & 41.1\% & 33.6$\times$ & 53.3\% & 41.1\% & 21.7$\times$ & 72.6\% & 18.3$\times$ & \textbf{58.5\%} \\
 & \cmark & \xmark & 19.9$\times$ & \textbf{57.9\%} & \textbf{43.9\%} & 26.4$\times$ & \textbf{65.8\%} & \textbf{54.5\%} & 19.7$\times$ & 63.2\% & 17.2$\times$ & 52.1\% \\
 & \xmark & \cmark & 23.3$\times$ & 49.1\% & 34.7\% & 31.0$\times$ & 51.3\% & 38.0\% & 18.5$\times$ & \textbf{75.1\%} & 18.4$\times$ & 55.3\% \\
 & \xmark & \xmark & 21.4$\times$ & 46.2\% & 30.9\% & 31.2$\times$ & 47.9\% & 36.5\% & 16.7$\times$ & 63.6\% & 15.7$\times$ & 54.5\% \\
\midrule
\multirow{4}{*}{Gemma-4-31B-it} & \cmark & \cmark & 25.9$\times$ & 60.3\% & 46.7\% & 30.0$\times$ & 80.8\% & 64.7\% & 22.5$\times$ & \textbf{74.5\%} & 18.4$\times$ & \textbf{69.0\%} \\
 & \cmark & \xmark & 21.5$\times$ & \textbf{71.2\%} & \textbf{57.6\%} & 27.0$\times$ & \textbf{85.2\%} & \textbf{71.2\%} & 20.7$\times$ & 69.9\% & 17.7$\times$ & 65.8\% \\
 & \xmark & \cmark & 24.8$\times$ & 52.8\% & 36.1\% & 31.1$\times$ & 63.0\% & 47.4\% & 17.2$\times$ & 74.1\% & 17.6$\times$ & 62.8\% \\
 & \xmark & \xmark & 22.8$\times$ & 55.6\% & 40.8\% & 34.2$\times$ & 61.5\% & 45.4\% & 17.4$\times$ & 67.4\% & 17.7$\times$ & 62.6\% \\
\bottomrule
\end{tabular}

\caption{Results of Gemma-4 models with and without ICL and thinking
under the default compression budget. For each model, bold marks the
best configuration on each preservation metric. The configuration with
ICL and without thinking is the one used in the main experiments
(Table~\ref{tab:main_results}).}
\label{tab:thinking}
\end{table}

\label{app:thinking_effect}

We ran a 2$\times$2 ablation on the four Gemma-4 models, crossing ICL
demonstrations with thinking under identical prompts, compression
budgets, and decoding parameters (Table~\ref{tab:thinking}). ICL
without thinking, the configuration used in the main experiments,
performed best on the parallel and conditional settings for
Gemma-4-12B-it and Gemma-4-31B-it and yielded the highest overall
All-Rec among all configurations (66.8\% for Gemma-4-31B-it). For these
two models, thinking improved the imbalanced setting with or without
ICL but lowered parallel and conditional performance when combined with
ICL. The two smaller models behaved differently: without ICL, thinking
improved all four All-Rec metrics, and this configuration outperformed
the main one on all four for Gemma-4-E2B-it and on three of four for
Gemma-4-E4B-it. Because no single configuration was best for every
model and setting, we used ICL without thinking, which gave the
strongest models their best results, for all general LLMs.

\subsection{Compressor Configuration}
\label{app:compression_prompt_reference}

All compressors except RECOMP (abstractive) received a sentence budget
$K$. Under the default budget, $K=2h$ for parallel and conditional cases
with $h$ hops and $K=2n+1$ for imbalanced cases with $n$ perspectives,
where the extra sentence leaves room to state that the question is
open-ended. The 0.5$\times$ and 2$\times$ budgets in
Section~\ref{sec:budget_analysis} halved or doubled the hop or perspective term.
RECOMP (abstractive) has no length control and was run once with its
default settings.

General LLMs received a system instruction, ICL demonstrations, and a
final user message containing the question, the original context, and
$K$. The demonstrations were built with the same pipeline as the
benchmark and excluded from evaluation. Each demonstration followed the
budget of its own case, so one prompt could contain several values of
$K$, while the final message stated the budget of the test case.
Thinking was disabled in the main experiments. Open-weight models ran on vLLM in bfloat16 with
temperature 0 and at most 65,536 generated tokens; the proprietary models
used temperature 0, at most 1,024 output tokens, and no thinking or
reasoning. The complete prompts are provided in Section~\ref{app:compression_prompts}.

The dedicated compressors used their released checkpoints. RECOMP
(extractive) and EXIT kept the $K$ highest scoring sentences, with EXIT
scoring each sentence within its own document, and BRIEF-Pro received
$K$ through its standard prompt. RECOMP (abstractive), trained on inputs
of 512 tokens, received up to 8,192 input tokens so that every complete
context fit.

\subsection{Compression Case Study}
\label{app:case_study}

We present an illustrative two-hop parallel case that contrasts
complete preservation with common failure patterns. Under the default
budget, this case allows four sentences, enough to state each target
proposition in its own sentence. RECOMP (abstractive), which has no
length control, is omitted. We chose this case because only one of the
remaining sixteen compressors preserves all four target propositions,
while several others retain both answers but no longer distinguish the
two series.

For readability, the input box below shows only the four supporting
paragraphs and one related distractor paragraph; ellipses mark omitted
text. Table~\ref{tab:case_study} reports the judgements for all sixteen
compressors, and we show the outputs of seven that illustrate complete
preservation and distinct failure patterns, reproduced verbatim apart
from line wrapping and the typesetting of quotation marks and Markdown
emphasis.

\begin{promptbox}{Parallel Case Input}

\textbf{Record ID.}

\texttt{parallel\_2hop\_0241}

\textbf{Question.}

In what county is the community of Lirion found, in the state where the
series Midnight Spotlight is set?

\textbf{Original context excerpt.}

\textbf{Document [6] (Title: The Midnight Spotlight).}

The drama series Midnight Spotlight, set in the fictional town of
Derby, Virelia (``The Hat Capital of the World''), featured a local
talk show aptly titled ``The Midnight Spotlight'', which was created by
housewife and talk television fan Jennifer Bass, which aired on cable
access. \ldots

\textbf{Document [4] (Title: Lirion (community), Virelia).}

Lirion is an unincorporated community located in the town of Lirion,
in Luminara County, Virelia. Lirion is located on Virelia Highway 36
west-southwest of Burlington. \ldots

\textbf{Document [17] (Title: The Midnight Spotlight).}

The documentary series Midnight Spotlight is set in Derby, Nocturnia, a
fictional town nicknamed ``The Hat Capital of the World'', and centers
on a cable-access talk show of the same name created by Jennifer Bass,
a housewife devoted to television. \ldots

\textbf{Document [10] (Title: Lirion (community), Nocturnia).}

Lirion is an unincorporated community in the town of Lirion, Ecliptica
County, Nocturnia, United States. It sits along Highway 36 in a rural
area west-southwest of Burlington. \ldots

\textbf{Distractor: Document [21] (Title: Lirion, New Jersey).}

Lirion is an unincorporated community located within Bernards Township,
in Somerset County, New Jersey, United States. \ldots

\textbf{Target evidence propositions.}

\textit{Branch 1: Midnight Spotlight (drama series)}

\begin{itemize}
    \setlength{\itemsep}{2pt}
    \item[$p_1$] The drama series Midnight Spotlight is set in Virelia.
    \item[$p_2$] Lirion (community) is located in Luminara County,
    Virelia.
\end{itemize}

\textit{Branch 2: Midnight Spotlight (documentary series)}

\begin{itemize}
    \setlength{\itemsep}{2pt}
    \item[$p_3$] The documentary series Midnight Spotlight is set in
    Nocturnia.
    \item[$p_4$] Lirion (community) is located in Ecliptica County,
    Nocturnia.
\end{itemize}

The two interpretations therefore yield the answers Luminara County and
Ecliptica County, respectively. Complete preservation requires
retaining both propositions in both branches.

\end{promptbox}

\begin{promptbox}{Claude-Haiku-4.5 Output}

The Midnight Spotlight series appears in the material as two records
linking a fictional town setting, a state, and a community named Lirion
with its county. One Midnight Spotlight paragraph says the drama series
is set in the fictional town of Derby, Virelia, and the Lirion
(community), Virelia paragraph identifies Lirion as an unincorporated
community located in Luminara County, Virelia. Another Midnight
Spotlight paragraph says the documentary series is set in the fictional
town of Derby, Nocturnia, and the Lirion (community), Nocturnia
paragraph identifies Lirion as an unincorporated community in Ecliptica
County, Nocturnia. Read together, those setting and county records
yield Luminara County on the Virelia side and Ecliptica County on the
Nocturnia side.

\end{promptbox}

\begin{table}[!h]
\centering
\setlength{\tabcolsep}{8pt}
\begin{tabular}{@{}lccccccc@{}}
\toprule
\textbf{Method} & $\mathbf{p_1}$ & $\mathbf{p_2}$ & $\mathbf{p_3}$ & $\mathbf{p_4}$
& \textbf{Retained} & \textbf{Para-Rec} & \textbf{All-Rec} \\
\midrule
BRIEF-Pro & \texttt{yes} & \texttt{no} & \texttt{no} & \texttt{yes} & 2/4 & \texttt{no} & \texttt{no} \\
RECOMP (extractive) & \texttt{no} & \texttt{yes} & \texttt{no} & \texttt{yes} & 2/4 & \texttt{no} & \texttt{no} \\
EXIT & \texttt{no} & \texttt{no} & \texttt{yes} & \texttt{yes} & 2/4 & \texttt{no} & \texttt{no} \\
\midrule
Llama-3.2-1B-Instruct & \texttt{no} & \texttt{no} & \texttt{no} & \texttt{no} & 0/4 & \texttt{no} & \texttt{no} \\
Llama-3.2-3B-Instruct & \texttt{no} & \texttt{yes} & \texttt{no} & \texttt{yes} & 2/4 & \texttt{no} & \texttt{no} \\
Llama-3.1-8B-Instruct & \texttt{no} & \texttt{yes} & \texttt{no} & \texttt{no} & 1/4 & \texttt{no} & \texttt{no} \\
Qwen3-4B & \texttt{no} & \texttt{yes} & \texttt{no} & \texttt{yes} & 2/4 & \texttt{no} & \texttt{no} \\
Qwen3-8B & \texttt{no} & \texttt{yes} & \texttt{no} & \texttt{yes} & 2/4 & \texttt{no} & \texttt{no} \\
Qwen3-14B & \texttt{no} & \texttt{yes} & \texttt{no} & \texttt{yes} & 2/4 & \texttt{no} & \texttt{no} \\
Qwen3-32B & \texttt{no} & \texttt{yes} & \texttt{no} & \texttt{yes} & 2/4 & \texttt{no} & \texttt{no} \\
Gemma-4-E2B-it & \texttt{yes} & \texttt{yes} & \texttt{no} & \texttt{yes} & 3/4 & \texttt{no} & \texttt{no} \\
Gemma-4-E4B-it & \texttt{no} & \texttt{yes} & \texttt{no} & \texttt{yes} & 2/4 & \texttt{no} & \texttt{no} \\
Gemma-4-12B-it & \texttt{no} & \texttt{yes} & \texttt{no} & \texttt{yes} & 2/4 & \texttt{no} & \texttt{no} \\
Gemma-4-31B-it & \texttt{no} & \texttt{yes} & \texttt{no} & \texttt{yes} & 2/4 & \texttt{no} & \texttt{no} \\
\midrule
GPT-5.4-mini & \texttt{no} & \texttt{yes} & \texttt{no} & \texttt{yes} & 2/4 & \texttt{no} & \texttt{no} \\
Claude-Haiku-4.5 & \texttt{yes} & \texttt{yes} & \texttt{yes} & \texttt{yes} & 4/4 & \texttt{yes} & \texttt{yes} \\
\bottomrule
\end{tabular}
\caption{Proposition-level judgements for the parallel case study under
the default budget. A \texttt{yes} indicates that the corresponding
proposition is preserved. Para-Rec is positive when both $p_1$ and
$p_3$, which distinguish the two candidate referents, are preserved;
All-Rec is positive only when all four propositions are preserved.}
\label{tab:case_study}
\end{table}

\begin{promptbox}{Gemma-4-31B-it Output}

Midnight Spotlight is described as being set in the state of Virelia in
one record and in the state of Nocturnia in another. In Virelia, the
community of Lirion is located in Luminara County. In Nocturnia, the
community of Lirion is located in Ecliptica County.

\end{promptbox}

\begin{promptbox}{Qwen3-14B Output}

The community of Lirion is mentioned in multiple locations: one in
Luminara County, Virelia; another in Ecliptica County, Nocturnia; and a
third in Somerset County, New Jersey. The series \textit{Midnight
Spotlight} is set in the fictional town of Derby, Virelia, which is in
the same state as the Lirion in Luminara County. Therefore, the Lirion
in Luminara County, Virelia, is the most relevant to the question. The
county is Luminara County.

\end{promptbox}

\begin{promptbox}{Gemma-4-E2B-it Output}

The series Midnight Spotlight is set in the fictional town of Derby,
Virelia, which is the setting for the drama. The community of Lirion is
an unincorporated community located in the town of Lirion, in Luminara
County, Virelia. The Cape Lirion document places Cape Lirion in the
Northwest Territories, Canada. The Lirion (community), Nocturnia
document places Lirion in Ecliptica County, Nocturnia, United States.

\end{promptbox}

\begin{promptbox}{Llama-3.2-3B-Instruct Output}

The community of Lirion is found in the state of Virelia, which is set
in the fictional town of Derby, Nocturnia, in the series Midnight
Spotlight. The Lirion in the series is an unincorporated community in
the town of Lirion, Ecliptica County, Nocturnia, United States, with a
post office using ZIP code 53148. Another Lirion is an unincorporated
community in the town of Lirion, Luminara County, Virelia, United
States, with a post office using ZIP code 53148. Read together, those
Lirion records identify the Lirion in the series as the one in
Nocturnia and the Lirion in Virelia as the one in Luminara County.

\end{promptbox}

\begin{promptbox}{EXIT Output}

Lirion (community), Nocturnia \\
Lirion is an unincorporated community in the town of Lirion, Ecliptica
County, Nocturnia, United States. Lirion, Oregon \\
Lirion is a city in Linn County, Oregon, United States. The Midnight
Spotlight \\
The documentary series Midnight Spotlight is set in Derby, Nocturnia, a
fictional town nicknamed ``The Hat Capital of the World'', and centers
on a cable-access talk show of the same name created by Jennifer Bass,
a housewife devoted to television. Lirion, New Jersey \\
Lirion is an unincorporated community located within Bernards
Township, in Somerset County, New Jersey, United States.

\end{promptbox}

\begin{promptbox}{RECOMP (extractive) Output}

Lirion (community), Virelia Lirion is an unincorporated community
located in the town of Lirion, in Luminara County, Virelia. Lirion
(community), Nocturnia Lirion is an unincorporated community in the
town of Lirion, Ecliptica County, Nocturnia, United States. Lirion,
Oregon Lirion is a city in Linn County, Oregon, United States. Lirion,
New Jersey Lirion is an unincorporated community located within
Bernards Township, in Somerset County, New Jersey, United States.

\end{promptbox}

\paragraph{Observation.}
Claude-Haiku-4.5 is the only compressor that preserves all four
propositions and the only one that achieves Para-Rec. It does so by
keeping the genre labels that distinguish the two series, tying the
drama series to Virelia and the documentary series to Nocturnia, each
paired with its own Lirion. The other compressors lose mainly $p_1$ and
$p_3$, which only three and two compressors preserve, respectively,
whereas $p_2$ and $p_4$ are each preserved by at least thirteen.
Gemma-4-31B-it shows the most consequential failure: it keeps both
states and both counties, so both gold answers remain recoverable, but
by dropping the genre labels it describes a single series with two
conflicting settings rather than two distinct series. Gemma-4-12B-it,
Qwen3-32B, and GPT-5.4-mini produce the same merged account. Qwen3-14B
lists both counties but resolves the question to the Virelia reading
alone. Gemma-4-E2B-it attaches the drama label to the Virelia setting,
but it omits the documentary setting and spends a sentence on the
unrelated Cape Lirion. Llama-3.2-3B-Instruct cross-binds the two
branches, placing Lirion in Virelia yet describing Virelia as set in
Derby, Nocturnia. The extractive methods favor sentences that mention
Lirion: EXIT keeps the Nocturnia branch but nothing from the Virelia
branch and spends two of its four sentences on the Lirion communities
in Oregon and New Jersey, while RECOMP (extractive) fills all four
sentences with Lirion county statements, including the same two
distractors, and omits both series. Most compressors thus retain the
answers but not the distinction that makes the question ambiguous, a
loss that answer accuracy alone would not reveal in this case.

\subsection{Inference FLOPs Estimation}
\label{app:flops_estimation}

We estimated major matrix multiplication FLOPs from each method's public
architecture using its actual sequence lengths.
Each multiply-add counts as two FLOPs.
We included attention and feed-forward projections, attention matrix
multiplications, cross-attention where applicable, and vocabulary
projections.
We assumed KV caching and excluded non-matrix operations, batching,
padding, preprocessing, and judge computation.
These estimates represent architectural computation rather than latency
or energy consumption.
Proprietary API models were excluded.

\paragraph{Decoder-only compressors.}
Let \(P\) and \(G\) denote the complete prompt length and compressed
output length under the corresponding family tokenizer.
With KV caching, the Transformer processes
\begin{equation}
T=P+G-1
\end{equation}
positions, while the vocabulary projection is evaluated at \(Z=G\)
positions.
For full attention and sliding-window attention with window \(w\), we
define
\begin{align}
\mathcal{A}_{\infty}(T)
&=\frac{T(T+1)}{2},\\
\mathcal{A}_{w}(T)
&=
\begin{cases}
\mathcal{A}_{\infty}(T), & T\leq w,\\[3pt]
wT-\dfrac{w(w-1)}{2}, & T>w.
\end{cases}
\end{align}

Qwen3, Llama, and BRIEF-Pro follow
\(F_m(T,Z)=a_mT+b_m\mathcal{A}_{\infty}(T)+c_mZ\), while Gemma-4
separates local and global attention.
One representative substitution from each model family is
\begin{align}
F_{\text{Qwen3-4B}}
&=
7{,}266{,}631{,}680T
+589{,}824\mathcal{A}_{\infty}(T)
\nonumber\\
&\quad
+777{,}912{,}320Z,
\\[3pt]
F_{\text{Llama-3.1-8B}}
&=
13{,}958{,}643{,}712T
+524{,}288\mathcal{A}_{\infty}(T)
\nonumber\\
&\quad
+1{,}050{,}673{,}152Z,
\\[3pt]
F_{\text{Gemma-4-E4B}}
&=
7{,}945{,}584{,}640T
+286{,}720\mathcal{A}_{512}(T)
\nonumber\\
&\quad
+114{,}688\mathcal{A}_{\infty}(T)
+1{,}342{,}177{,}280Z.
\end{align}
All other evaluated sizes used their own architecture coefficients in
our calculation scripts rather than estimates scaled from these
examples.
For BRIEF-Pro, we used the coefficients of its released
Llama-3.2-3B-Instruct backbone.

\paragraph{Prefix caching.}
For each open-weight LLM, we constructed one prefix cache for each of the
parallel, conditional, imbalanced science, and imbalanced policy
subsets.
Let \(p_{m,s}\) be the length of the longest common token prefix shared
by the \(n_s\) serialized prompts in subset \(s\).
For case \(i\), the prefix-amortized computation is
\begin{align}
\widetilde{F}_{m,i}
&=
F_m(T_{m,i},G_{m,i})
-F_m(p_{m,s},0)
\nonumber\\
&\quad
+\frac{F_m(p_{m,s},0)}{n_s}.
\end{align}
The second term removes the cached prefix computation from each request,
while the final term distributes its one-time prefill cost across all
cases in that subset.

\paragraph{Dedicated compressors.}
Let \(M\) denote the number of candidate sentences, \(\ell_0\) the
question length, \(\ell_j\) each candidate length, \(n\) and \(r\) the
RECOMP abstractive encoder and decoder lengths, and \(p_j\) each EXIT
classification-prompt length.
Using the released execution paths,
\begin{align}
F_{\text{RECOMP-ext}}
&=
\sum_{j=0}^{M}
\left(
169{,}869{,}312\ell_j
+36{,}864\ell_j^2
\right)
\nonumber\\
&\quad
+1{,}536M,
\\[3pt]
F_{\text{RECOMP-abs}}
&=
704{,}643{,}072n
+98{,}304n^2
+770{,}441{,}216r
\nonumber\\
&\quad
+98{,}304
\left(
\mathcal{A}_{\infty}(r)+nr
\right),
\\[3pt]
F_{\text{EXIT}}
&=
\sum_{j=1}^{M}
\left[
5{,}169{,}086{,}464p_j
+147{,}456\mathcal{A}_{\infty}(p_j)
\right].
\end{align}
The RECOMP abstractive expression uses its default beam width of one.
The EXIT expression includes its unmerged rank-64 LoRA adapter and
full-sequence vocabulary projection.
Dedicated compressors did not use ICL and therefore received no prefix
cache amortization.

\paragraph{Aggregation.}
For methods without prefix caching, we set
\(\widetilde{F}_{m,i}=F_{m,i}\).
We then averaged over all \(N=4{,}102\) cases and expressed the result in
TFLOPs:
\begin{equation}
\overline{F}_m
=
\frac{1}{10^{12}N}
\sum_{i=1}^{N}\widetilde{F}_{m,i}.
\end{equation}
We calculated FLOPs for each case before averaging because attention
cost is nonlinear in sequence length.
Table~\ref{tab:flops_per_method} reports $\overline{F}_m$ for each
method, together with the overall All-Rec $A_m$ used in EPR.

\begin{table}[!h]
\centering
\small
\begin{tabular}{lrr}
\toprule
Method & $A_m$ (\%) & $\overline{F}_m$ (TFLOPs) \\
\midrule
\multicolumn{3}{l}{\textit{Dedicated compressors}} \\
RECOMP (abstractive) & 0.5 & 6.21 \\
BRIEF-Pro & 6.7 & 21.94 \\
RECOMP (extractive) & 4.8 & 1.49 \\
EXIT & 6.3 & 59.96 \\
\midrule
\multicolumn{3}{l}{\textit{Open-weight LLMs}} \\
Llama-3.2-1B-Instruct & 0.3 & 13.74 \\
Llama-3.2-3B-Instruct & 4.9 & 37.47 \\
Llama-3.1-8B-Instruct & 18.6 & 75.53 \\
Qwen3-4B & 19.8 & 60.04 \\
Qwen3-8B & 24.7 & 82.17 \\
Qwen3-14B & 33.8 & 139.49 \\
Qwen3-32B & 39.9 & 338.80 \\
Gemma-4-E2B-it & 5.5 & 19.27 \\
Gemma-4-E4B-it & 25.0 & 33.82 \\
Gemma-4-12B-it & 53.8 & 90.66 \\
Gemma-4-31B-it & 66.8 & 239.60 \\
\bottomrule
\end{tabular}
\caption{Overall All-Rec $A_m$ and average inference computation
$\overline{F}_m$ per case under the default compression budget
(RECOMP (abstractive) without length control), with prefix amortization
for methods using ICL. These two quantities give the EPR values in
Table~\ref{tab:main_results}. Proprietary API models are excluded.}
\label{tab:flops_per_method}
\end{table}

\subsection{Results across Output Budgets}
\label{app:budget_results}
\begin{table}[!h]
\centering
\small
\setlength{\tabcolsep}{3.5pt}

\begin{tabular}{@{}l ccc ccc ccc@{}}
\toprule
\multirow{2}{*}{\textbf{Method}}
& \multicolumn{3}{c}{\textbf{Parallel}}
& \multicolumn{3}{c}{\textbf{Conditional}}
& \multicolumn{3}{c}{\textbf{Imbalanced}} \\
\cmidrule(lr){2-4} \cmidrule(lr){5-7} \cmidrule(l){8-10}
& 0.5$\times$ & 1$\times$ & 2$\times$ & 0.5$\times$ & 1$\times$ & 2$\times$ & 0.5$\times$ & 1$\times$ & 2$\times$ \\
\midrule
BRIEF-Pro & 0.0 (133) & 0.0 (202) & 0.9 (332) & 8.2 (143) & 19.9 (201) & 63.9 (332) & 0.2 (186) & 0.2 (327) & 1.4 (586) \\
RECOMP (extractive) & 0.0 (111) & 1.4 (219) & 6.6 (431) & 0.5 (104) & 2.5 (207) & 11.6 (410) & 1.5 (169) & 8.1 (308) & 22.1 (587) \\
EXIT & 0.1 (109) & 3.0 (217) & 14.9 (430) & 1.1 (99) & 4.5 (200) & 16.3 (402) & 2.8 (191) & 9.2 (342) & 21.2 (633) \\
\midrule
Llama-3.2-1B-Instruct & 0.0 (106) & 0.0 (171) & 0.0 (306) & 0.2 (66) & 0.1 (120) & 2.4 (237) & 0.2 (147) & 0.6 (230) & 1.6 (353) \\
Llama-3.2-3B-Instruct & 3.0 (119) & 2.8 (202) & 2.9 (331) & 3.7 (72) & 6.7 (124) & 12.7 (232) & 4.0 (112) & 4.7 (173) & 8.1 (274) \\
Llama-3.1-8B-Instruct & 11.4 (102) & 13.0 (211) & 24.7 (388) & 8.9 (62) & 23.5 (122) & 38.8 (237) & 14.7 (125) & 17.8 (210) & 24.0 (364) \\
Qwen3-4B & 14.2 (117) & 14.4 (211) & 18.8 (305) & 13.3 (63) & 26.6 (136) & 44.9 (242) & 12.1 (126) & 17.4 (214) & 22.0 (332) \\
Qwen3-8B & 13.8 (87) & 23.8 (150) & 26.8 (247) & 14.0 (57) & 29.3 (110) & 47.4 (186) & 18.7 (114) & 21.8 (184) & 31.0 (290) \\
Qwen3-14B & 21.6 (107) & 33.0 (189) & 37.3 (272) & 21.8 (61) & 43.1 (119) & 61.0 (194) & 19.5 (115) & 27.3 (203) & 38.0 (329) \\
Qwen3-32B & 24.2 (99) & 42.1 (184) & 50.0 (277) & 20.5 (55) & 44.6 (115) & 64.1 (192) & 22.1 (121) & 35.4 (220) & 47.1 (334) \\
Gemma-4-E2B-it & 0.8 (75) & 2.3 (119) & 2.4 (181) & 2.6 (41) & 8.3 (83) & 13.9 (139) & 5.2 (95) & 5.1 (141) & 10.6 (222) \\
Gemma-4-E4B-it & 13.9 (97) & 24.6 (157) & 29.3 (256) & 13.1 (58) & 27.9 (105) & 48.6 (179) & 14.2 (124) & 23.1 (180) & 32.9 (283) \\
Gemma-4-12B-it & 25.7 (89) & 43.9 (141) & 59.3 (243) & 23.6 (52) & 54.5 (96) & 72.7 (170) & 40.3 (130) & 58.2 (194) & 68.6 (277) \\
Gemma-4-31B-it & 33.3 (87) & 57.6 (130) & 71.4 (203) & 47.5 (55) & 71.2 (94) & 85.9 (166) & 55.0 (122) & 68.1 (186) & 75.8 (256) \\
\midrule
GPT-5.4-mini & 19.3 (110) & 31.9 (163) & 39.4 (242) & 23.8 (56) & 37.0 (102) & 52.1 (173) & 34.7 (131) & 43.3 (187) & 52.0 (267) \\
Claude-Haiku-4.5 & 50.1 (109) & 63.8 (225) & 70.0 (338) & 41.1 (59) & 70.0 (122) & 88.3 (251) & 43.5 (112) & 54.3 (168) & 63.3 (279) \\
\bottomrule
\end{tabular}

\caption{All-Rec in percent, with the average output length in tokens
in parentheses, at 0.5$\times$, 1$\times$, and 2$\times$ the default sentence
budget. The 1$\times$ columns correspond to Table~\ref{tab:main_results};
imbalanced results pool the science and policy cases.}
\label{tab:budget_results}
\end{table}

Table~\ref{tab:budget_results} reports the full results of the budget
analysis in Section~\ref{sec:budget_analysis} for all methods with explicit
length control. 
The three budgets correspond to 0.5$\times$,
1$\times$, and 2$\times$ the default sentence budget described in
Section~\ref{app:compression_prompt_reference}. RECOMP (abstractive)
was excluded because it did not reliably follow length constraints.
For each method and budget, we report All-Rec and, in parentheses, the
average output length in tokens, averaged over the cases of each setting.

\clearpage

\section{Configuration and Validation of LLM-as-Judge}
\label{app:judge_configuration}

\subsection{Judge Configuration}

We used \texttt{GPT-5.2} as the automatic judge, with reasoning
effort set to \texttt{high}. Each compressor output was evaluated
independently against the corresponding target evidence propositions.
We used one model response per judgement, without repeated sampling or
majority voting. Parameters not explicitly specified in our
implementation used the API defaults.

\subsection{Decision Procedure}

For the parallel setting, the judge received the compressed context and
one target proposition. When a proposition involved an ambiguous entity,
we additionally provided metadata identifying the competing unambiguous
entities. The judge then accepted the proposition only if the statement
supporting it was tied to the correct entity through that entity's
distinguishing qualifier or an unambiguous equivalent. The exact label
was not required, and the context could also keep the alternatives
apart, each with its own qualifier, provided that the facts of one were
not attributed to the other. Evidence about another listed entity, or
evidence retaining only the bare ambiguous name, could therefore not
support the proposition.

The conditional setting used two evaluation procedures.
Ordinary propositions were evaluated with the general proposition-level
entailment prompt.
For propositions governed by an explicit condition, the judge jointly
returned \texttt{condition\_judgement} and \texttt{overall\_judgement}.
We used \texttt{condition\_judgement} as the preservation label for the
condition-bearing proposition and \texttt{overall\_judgement} as the
label for the associated general proposition. The latter was evaluated
under the scope imposed by the condition and had to be \texttt{no} when
the condition itself was not established by the compressed context. The
accompanying disambiguation proposition describes the broader
ambiguity resolved by the condition and helps the judge interpret its
scope; it is not itself an evaluated proposition.

For the science and policy subsets of the imbalanced setting, the judge
received the benchmark question, the compressed context, and one target
perspective. It determined whether the compressed context preserved
that perspective as an answer to the question. For policy perspectives,
the compressed context also had to identify the jurisdiction to which
the answer applies.

All model responses were requested in JSON format. The standard
procedures returned \texttt{\{"judgement":"yes"\}} or
\texttt{\{"judgement":"no"\}}, whereas the condition-aware procedure
returned both \texttt{condition\_judgement} and
\texttt{overall\_judgement}. Failed or unparseable requests were
retried; empty outputs and, in the imbalanced setting, the placeholder
output ``This passage doesn't contain relevant information to the
question.'' were labeled \texttt{no} without a judge call. These labels
were aggregated into the metrics of Section~\ref{sec:experiments}; the
complete prompts are given in Section~\ref{app:evaluation_prompts}.

\subsection{Automatic Judge Validation}
\label{app:judge_validation}
\begin{table}[h]
\centering
\small
\begin{tabular}{@{}lrrr@{}}
\toprule
\textbf{Setting}
& \textbf{Decisions}
& \textbf{Exact Matches}
& \textbf{Agreement} \\
\midrule
Parallel   & 100 & 95  & 95.0\% \\
Conditional & 100 & 91 & 91.0\% \\
Imbalanced & 100 & 93  & 93.0\% \\
\midrule
Overall    & 300 & 279 & 93.0\% \\
\bottomrule
\end{tabular}
\caption{Decision-level exact agreement between the automatic judge and
manual validation judgements.}
\label{tab:judge_validation}
\end{table}

To assess the reliability of the automatic judge, we manually validated
100 sampled proposition-level decisions from each evidence setting, for
a total of 300 decisions.
Each decision corresponds to one judge call for one target proposition;
condition-aware calls in the conditional setting also assess the
accompanying condition and return both \texttt{condition\_judgement}
and \texttt{overall\_judgement}.
The samples were stratified by the judge's output to cover both
positive and negative decisions.

For the parallel setting, we sampled 50 decisions judged as \texttt{yes}
and 50 judged as \texttt{no}.
For the imbalanced setting, the sample contained 50 science and 50
policy decisions, while also comprising 50 \texttt{yes} and 50
\texttt{no} judgements.
For the conditional setting, we sampled 20 decisions from each of the
following five output categories: general \texttt{judgement=yes};
general \texttt{judgement=no}; \texttt{condition\_judgement=no} with
\texttt{overall\_judgement=no}; \texttt{condition\_judgement=yes} with
\texttt{overall\_judgement=yes}; and \texttt{condition\_judgement=yes}
with \texttt{overall\_judgement=no}.

We measured decision-level exact agreement between the automatic and
manual judgements.
For outputs containing a single judgement field, the decision was
counted as an agreement only when that field matched the manual
judgement.
For condition-aware outputs, both \texttt{condition\_judgement} and
\texttt{overall\_judgement} had to match their respective manual
judgements; a mismatch in either field caused the entire decision to be
counted as a disagreement.

As shown in Table~\ref{tab:judge_validation}, the automatic judge
achieves at least 91.0\% exact agreement in every setting and 93.0\%
overall.

\clearpage

\section{Prompts}
\label{app:prompts}

We collect all prompts used for compression, evaluation, and the downstream pilot study, together with abridged versions of three key data-construction prompts.
Values enclosed by double braces are inserted at runtime. 
The construction prompts are abridged to omit schema validation and formatting details.
For compression, we reproduce the system and final user prompt templates and indicate where the ICL cases are inserted.
All evaluation prompts are reproduced in full.

\subsection{Data Construction Prompts}
\label{app:construction_prompts}

\begin{promptbox}{Conditioned-Chain Construction Prompt (Abridged)}

\textbf{System prompt.}

You are a careful JSON rewriter. Return only valid JSON with the
required fields.

\textbf{User prompt.}

You are given a fictionalized multi-hop question, its logical chain,
and its answer.

Transform the example into a condition-dependent instance.

\textbf{Requirements:}

\begin{itemize}
    \setlength{\itemsep}{2pt}

    \item Introduce one specific condition, stored as
    \texttt{key}, that restricts exactly one proposition in the
    logical chain.

    \item Rewrite the affected proposition so that it is valid under
    the condition and cannot be interpreted as an unrestricted claim.

    \item Keep the remaining reasoning hops coherent with the
    conditioned proposition and preserve the original number of hops.

    \item Write a
    \texttt{disambiguation\_proposition} establishing that the
    relevant relation can differ across conditions and therefore
    requires the key for disambiguation.

    \item The condition must be both logically compatible with the
    question and necessary for identifying the intended answer.

    \item Return exactly
    \texttt{question}, \texttt{key},
    \texttt{disambiguation\_proposition},
    \texttt{logical\_chain}, and \texttt{answer}.
\end{itemize}

\texttt{\#\#\#\# Input}

\texttt{\{\{INPUT\_JSON\}\}}

\texttt{\#\#\#\# Output}

\end{promptbox}

\begin{promptbox}{Dominant-Proposition Retrieval Query Prompt (Abridged)}

\textbf{System prompt.}

You are a precise JSON editor. Output only JSON that conforms to the
schema.

\textbf{User prompt.}

You are given a benchmark question and one target evidence
proposition. Write one compact search query for retrieving documents
that substantively support that proposition.

For a scientific proposition, preserve the scientific topic, center
the target mechanism or explanation, and add a small number of
support-oriented retrieval terms.

For a policy proposition, preserve the jurisdiction, policy topic, and
any concrete rule, deadline, threshold, duty, right, or exception.
Prefer wording likely to retrieve official or authoritative sources.

Do not write a full-sentence question, introduce competing
propositions, or request a general comparison.

Output exactly:

\texttt{\{"rag\_input":"..."\}}

\texttt{\#\#\#\# Input}

\texttt{\{\{INPUT\_JSON\}\}}

\texttt{\#\#\#\# Output}

\end{promptbox}

\begin{promptbox}{Minority-Proposition Insertion Prompt (Abridged)}

\textbf{System prompt.}

You are a precise JSON editor. Output only JSON that conforms to the
schema.

\textbf{User prompt.}

You are given a question, a list of designated minority propositions,
and the paragraphs already selected for the dominant propositions.

For each minority proposition:

\begin{enumerate}
    \setlength{\itemsep}{2pt}

    \item Choose one topically compatible host paragraph.

    \item Insert exactly one natural sentence.

    \item Ensure that the inserted sentence itself entails the entire
    minority proposition.

    \item Preserve all answer-critical mechanisms, jurisdictions,
    values, deadlines, conditions, and exceptions.

    \item Keep the original paragraph as intact as possible and make
    only minimal local edits needed for coherence.

    \item Do not mention dataset labels, evidence roles, sources, or
    the insertion process.
\end{enumerate}

Return one insertion record per minority proposition, including its
index, the host paragraph identifier, the inserted sentence, and the
complete revised paragraph.

\texttt{\#\#\#\# Input}

\texttt{\{\{INPUT\_JSON\}\}}

\texttt{\#\#\#\# Output}

\end{promptbox}

\clearpage

\subsection{Compression Prompts}
\label{app:compression_prompts}  

\emph{[ICL cases]} marks where the ICL messages were inserted; the
prompting setup is described in
Section~\ref{app:compression_prompt_reference}.

\begin{promptbox}{Compression: Parallel Evidence}

\textbf{System prompt.}

Write a high-quality summary of the provided documents with respect to
the question.

The question may contain potentially ambiguous entities, and within the
documents some entities mentioned in the question may refer to multiple
different referents.

Preserve all such possible interpretations and supporting evidence in
the summary.

Include all relevant evidence for different possible referents,
preserve conflicting evidence when present, and prioritize and
front-load the most critical points while remaining concise.

\textbf{ICL cases.}

\textit{[ICL cases]}

\textbf{User prompt.}

\texttt{\#\#\#} This is the question:
\texttt{\{\{QUESTION\}\}}

\texttt{\#\#\#} These are the documents:

\texttt{\{\{ORIGINAL\_CONTEXT\}\}}

\texttt{\#\#\#} This is the summary (must include all relevant
evidence for different possible referents; prioritize and front-load
the most critical points; total length must not exceed
\texttt{\{\{K\}\}} sentences):

\end{promptbox}

\begin{promptbox}{Compression: Conditional Evidence}

\textbf{System prompt.}

Write a high-quality summary of the provided documents with respect to
the question.

The question may have different correct answers under different
conditions.

Preserve the condition-bearing evidence needed for an answer to be
correct, especially when a result holds only under a specific
condition.

Prioritize and front-load the most critical evidence needed to answer
the question, and do not omit the condition if the answer depends on
it.

\textbf{ICL cases.}

\textit{[ICL cases]}

\textbf{User prompt.}

\texttt{\#\#\#} This is the question:
\texttt{\{\{QUESTION\}\}}

\texttt{\#\#\#} These are the documents:

\texttt{\{\{ORIGINAL\_CONTEXT\}\}}

\texttt{\#\#\#} This is the summary (must preserve
condition-bearing evidence; prioritize and front-load the most critical
points; total length must not exceed \texttt{\{\{K\}\}} sentences):

\end{promptbox}
\clearpage

\begin{promptbox}{Compression: Imbalanced Evidence---Science}

\textbf{System prompt.}

Write a concise, document-grounded summary of the provided documents
with respect to the question.

The summary should stay focused on the question, preserve the answer
categories supported by the documents, and maximize coverage of
distinct perspectives when the documents support multiple answers.

Do not add facts that are not supported by the documents.

Prefer short, direct sentences that front-load the most important
evidence.

\textbf{ICL cases.}

\textit{[ICL cases]}

\textbf{User prompt.}

\texttt{\#\#\#} This is the question:
\texttt{\{\{QUESTION\}\}}

\texttt{\#\#\#} These are the documents:

\texttt{\{\{ORIGINAL\_CONTEXT\}\}}

\texttt{\#\#\#} This is the summary (must stay focused on the
question; must remain grounded in the documents; must maximize coverage
of distinct supported perspectives; total length must not exceed
\texttt{\{\{K\}\}} sentences):

\end{promptbox}

\begin{promptbox}{Compression: Imbalanced Evidence---Policy}

\textbf{System prompt.}

Write a concise, document-grounded summary of the provided documents
with respect to the question.

The summary should stay focused on the question, preserve the answer
categories supported by the documents, and maximize coverage of
distinct perspectives when the documents support multiple answers.

For policy questions, each answer category must clearly identify the
applicable country, region, legal system, or jurisdiction.

Do not add facts that are not supported by the documents.

Prefer short, direct sentences that front-load the most important
evidence.

\textbf{ICL cases.}

\textit{[ICL cases]}

\textbf{User prompt.}

\texttt{\#\#\#} This is the question:
\texttt{\{\{QUESTION\}\}}

\texttt{\#\#\#} These are the documents:

\texttt{\{\{ORIGINAL\_CONTEXT\}\}}

\texttt{\#\#\#} This is the summary (must stay focused on the
question; must remain grounded in the documents; must maximize coverage
of distinct supported perspectives; must identify the applicable
jurisdiction for each policy answer; total length must not exceed
\texttt{\{\{K\}\}} sentences):

\end{promptbox}

\clearpage

\subsection{Evaluation Prompts}
\label{app:evaluation_prompts}

Field names keep the spelling judgement used in our implementation. In
the parallel prompt, \texttt{\{\{CONSTRAINTS\_JSON\}\}} lists each
ambiguous entity with its unambiguous alternatives; ordinary conditional
propositions use this prompt without the disambiguation constraints and
rules. The policy prompt adds the three rules tagged
\textbf{[Policy only]} to the science prompt.

\begin{promptbox}{Evaluation: Parallel Evidence}

\textbf{System prompt.}

You are a precise JSON editor. Output only JSON that conforms to the
schema.

\textbf{User prompt.}

You are given a JSON input with two keys: \texttt{"context"} and
\texttt{"proposition"}.

Your task is to determine whether the proposition is entailed by the
context.

Entailment means that the proposition can be logically inferred from at
least one coherent part of the context.

\textbf{Important clarifications:}

\begin{itemize}
    \setlength{\itemsep}{2pt}
    \item The context may contain irrelevant, noisy, or even
    contradictory information. This does NOT invalidate entailment. As
    long as there exists at least one subset of the context from which
    the proposition can be reasonably inferred, the proposition should
    be considered entailed.
    \item The proposition does not need to match the wording used in
    the context. As long as the proposition holds under a reasonable
    interpretation of some part of the context, it should be considered
    entailed.
\end{itemize}

If the context entails the proposition, output exactly:

\texttt{\{"judgement":"yes"\}}

Otherwise, output exactly:

\texttt{\{"judgement":"no"\}}

Do not include any explanation or extra text. Output JSON only.

\textbf{Additional Disambiguation Constraints (all must be applied):}

\texttt{\{\{CONSTRAINTS\_JSON\}\}}

\textbf{Rules for judgement (apply to every constraint listed above):}

\begin{itemize}
    \setlength{\itemsep}{2pt}
    \item Each listed \texttt{"unambiguous entity"} consists of the
    ambiguous name plus a distinguishing qualifier (for example a
    country, a county, a date, a type, or an associated person or
    work). The constraint is about identifying the correct entity. It
    does not require the label's exact wording or its parenthetical
    form, and it does not require every part of the qualifier to be
    reproduced, as long as what appears is enough to single out that
    entity rather than the other listed alternatives.
    \item If the proposition mentions one of the listed
    \texttt{"unambiguous entities"}, then context can be judged as
    entailment only if the proposition-relevant statement in context is
    tied to that same entity, because that entity's distinguishing
    qualifier, or an unambiguous equivalent of it, appears in or is
    attached to that statement.
    \item If the proposition-relevant statement carries only the bare
    ambiguous name, with no distinguishing qualifier attached to it,
    the judgement must be \texttt{"no"}, even if that qualifier appears
    elsewhere in the context in connection with something else.
    \item If the proposition-relevant statement is tied to a different
    listed entity than the one the proposition mentions, the judgement
    must be \texttt{"no"} even if other content appears compatible.
    \item The context may instead keep the alternatives apart,
    attaching each one to its own distinguishing qualifier (for example
    ``X is a town in County A, and X is also a place in County B'').
    This satisfies the constraint for whichever alternative the
    proposition mentions, provided that alternative's facts are not
    attributed to the other.
    \item If several constraints are listed, the judgement must be
    \texttt{"no"} unless every constraint whose entities the
    proposition mentions is satisfied.
    \item These are necessary conditions only; all original entailment
    requirements must still be satisfied.
\end{itemize}

\texttt{\#\#\#\# Input}

\texttt{\{\{INPUT\_JSON\}\}}

\texttt{\#\#\#\# Output}

\end{promptbox}

\begin{promptbox}{Evaluation: Conditional Evidence---Condition-Aware}

\textbf{System prompt.}

You are a precise JSON editor. Output only JSON that conforms to the
schema.

\textbf{User prompt.}

You are given a JSON input with four keys: \texttt{"context"},
\texttt{"key"}, \texttt{"disambiguation\_proposition"}, and
\texttt{"proposition"}.

Your task is to make two entailment judgements.

The relationship among \texttt{"disambiguation\_proposition"},
\texttt{"key"}, and \texttt{"proposition"} is important:

\begin{itemize}
    \setlength{\itemsep}{2pt}
    \item \texttt{"disambiguation\_proposition"} describes the broader
    ambiguous set in which the condition from \texttt{"key"} selects a
    subset or imposes a constraint.
    \item Use \texttt{"disambiguation\_proposition"} to understand what
    kind of ambiguity the key is resolving and what collection of
    possible cases the key is restricting.
    \item This helps you judge whether the context entails the specific
    condition in \texttt{"key"}.
    \item \texttt{"key"} is a specific condition or scope restriction.
    \item \texttt{"proposition"} is a claim that must be interpreted
    under that specific condition.
    \item In other words, the proposition is not free-standing. Its
    later content must be understood as being restricted by the
    condition in \texttt{"key"}.
\end{itemize}

1. First, determine whether the condition in \texttt{"key"} is entailed
by the context. Write this as \texttt{"condition\_judgement"}. To judge
this, check whether the context contains or establishes the specific
conditioned scope expressed by the key. For example:

\begin{itemize}
    \setlength{\itemsep}{2pt}
    \item If the key is ``For the Lyra surnamed Voss'', then the context
    should contain or establish that specific Lyra surnamed Voss.
    \item If the key is ``For the original mechanism'', then the context
    should contain or establish that specific original mechanism.
\end{itemize}

If the context does not establish the condition itself, then
\texttt{"condition\_judgement"} should be \texttt{"no"}.

2. Then determine whether the proposition is entailed by the context,
but this second judgement must respect the condition:

\begin{itemize}
    \setlength{\itemsep}{2pt}
    \item If the context does not entail the condition in
    \texttt{"key"}, then \texttt{"overall\_judgement"} must be
    \texttt{"no"}.
    \item If the context does entail the condition in \texttt{"key"},
    then judge whether the full proposition is entailed by the context
    under that condition.
    \item The proposition should count as entailed only if the context
    supports the later information in the proposition as being about
    the scope restricted by the key.
    \item In other words, the context must not merely support a similar
    statement somewhere else; it must support that statement
    specifically under the condition introduced by the key.
\end{itemize}

Entailment means that the statement can be logically inferred from at
least one coherent part of the context.

\textbf{Important clarifications:}

\begin{itemize}
    \setlength{\itemsep}{2pt}
    \item The context may contain irrelevant, noisy, or even
    contradictory information. This does NOT invalidate entailment. As
    long as there exists at least one subset of the context from which
    the target statement can be reasonably inferred, it should be
    considered entailed.
    \item The key and proposition do not need to match the wording used
    in the context. As long as the context expresses the same meaning,
    they should be considered entailed.
\end{itemize}

Output exactly one JSON object in the following format:

\texttt{\{"condition\_judgement":"yes" or "no","overall\_judgement":"yes" or "no"\}}

Do not include any explanation or extra text. Output JSON only.

\texttt{\#\#\#\# Input}

\texttt{\{\{INPUT\_JSON\}\}}

\texttt{\#\#\#\# Output}

\end{promptbox}

\begin{promptbox}{Evaluation: Imbalanced Evidence}

\textbf{System prompt.}

You are a precise JSON judge. Output only JSON that conforms to the
schema.

\textbf{User prompt.}

You are given a JSON input with three keys: \texttt{"question"},
\texttt{"compressed\_context"}, and \texttt{"perspective"}.

Your task is to determine whether the \texttt{compressed\_context}
entails the perspective as an answer to the question.

Judge entailment by meaning, not by exact wording. The
\texttt{compressed\_context} does NOT need to restate every word or
every minor detail of the perspective. However, it must clearly express
the same answer in substance.

\textbf{Important clarifications:}

\begin{itemize}
    \setlength{\itemsep}{2pt}
    \item The \texttt{compressed\_context} may contain multiple different
    perspectives or answer candidates. This is allowed.
    \item If any coherent part of the \texttt{compressed\_context} points
    to the same underlying answer to the question as the perspective,
    judge \texttt{"yes"}.
    \item The judgement is about whether the perspective itself is
    entailed, not whether a broader topic related to it is mentioned.
    \item \textbf{[Policy only]} The jurisdiction matters. The \texttt{compressed\_context}
    must clearly indicate the country, region, legal system, or
    jurisdiction to which the entailed answer applies.
    \item \textbf{[Policy only]} If the \texttt{compressed\_context} states the substantive
    rule but does not identify the applicable jurisdiction, judge
    \texttt{"no"}.
    \item \textbf{[Policy only]} The jurisdiction does not need to be named with the exact
    same wording as the perspective. An unambiguous equivalent name,
    abbreviation, governing body, agency, or legal framework can count.
    \item A more general category does NOT count unless it clearly
    conveys the same answer as the perspective.
    \item Missing minor wording details can be acceptable, but some
    coherent part of the \texttt{compressed\_context} must still point
    to the same underlying answer to the question.
    \item If the \texttt{compressed\_context} merely mentions a broader
    family that could include the perspective, but does not itself
    point to the same underlying answer, judge \texttt{"no"}.
    \item If some coherent part of the \texttt{compressed\_context}
    clearly conveys the same answer as the perspective, even with
    paraphrasing, judge \texttt{"yes"}.
\end{itemize}

If the \texttt{compressed\_context} entails the perspective, output
exactly:

\texttt{\{"judgement":"yes"\}}

Otherwise, output exactly:

\texttt{\{"judgement":"no"\}}

Do not include any explanation or extra text. Output JSON only.

\texttt{\#\#\#\# Input}

\texttt{\{\{INPUT\_JSON\}\}}

\texttt{\#\#\#\# Output}

\end{promptbox}

\clearpage

\subsection{Downstream Pilot Prompts}
\label{app:downstream_pilot_prompts}

\begin{promptbox}{Downstream Reader Prompt}

\textbf{System prompt.}

You are a precise JSON-only question-answering reader.

\textbf{User prompt.}

You are a careful question-answering reader.

You will be given a JSON object with two fields:
\texttt{"query"}, the question to answer, and
\texttt{"context"}, the only context you may use.

Answer the query using only the context.

\textbf{Important rules:}

\begin{itemize}
    \setlength{\itemsep}{2pt}

    \item The query may be ambiguous because an entity mentioned in it
    can refer to multiple different referents.

    \item If the context supports multiple valid answers under
    different interpretations, include all distinct answers.

    \item If the context supports only one answer, include only that
    answer.

    \item Keep answers concise; do not add explanation unless it is
    needed to distinguish multiple answers.

    \item If the context does not contain enough evidence to answer,
    use \texttt{"insufficient\_information"}.
\end{itemize}

Output exactly one JSON object in this schema:

\texttt{\{"answers":["..."]\}}

Do not output markdown or extra text.

\texttt{\#\#\#\# Input}

\texttt{\{\{INPUT\_JSON\}\}}

\texttt{\#\#\#\# Output}

\end{promptbox}

\begin{promptbox}{Semantic Answer Evaluation Prompt}

\textbf{System prompt.}

You are a precise JSON-only semantic answer evaluator.

\textbf{User prompt.}

You are evaluating an answer to an ambiguous question.

The gold answer contains two valid answers, \texttt{answer\_1} and
\texttt{answer\_2}. A prediction is correct only if it semantically
includes BOTH gold answers. Exact wording is not required.

\textbf{Rules:}

\begin{itemize}
    \setlength{\itemsep}{2pt}

    \item Treat minor wording differences, punctuation,
    capitalization, and ordering as irrelevant.

    \item Treat equivalent units, names, and paraphrases as correct
    when the meaning is the same.

    \item If the prediction only contains one of the two gold answers,
    it is incorrect.

    \item If the prediction gives an answer that contradicts either
    gold answer or omits one gold answer, it is incorrect.

    \item Ignore extra explanation if both gold answers are clearly
    present.
\end{itemize}

Output exactly one JSON object:

\texttt{\{"correct":true/false,}\\
\texttt{"matched\_answer\_1":true/false,}\\
\texttt{"matched\_answer\_2":true/false\}}

\texttt{\#\#\#\# Input}

\texttt{\{\{INPUT\_JSON\}\}}

\texttt{\#\#\#\# Output}

\end{promptbox}

\end{document}